\documentclass[letterpaper, 10 pt, journal, twoside]{ieeetran} 

\IEEEoverridecommandlockouts                              

\usepackage{amsmath,amsfonts}
\usepackage{algorithmic}
\usepackage{algorithm}
\usepackage{array}
\usepackage[caption=false,font=normalsize]{subfig}
\usepackage{textcomp}
\usepackage{stfloats}
\usepackage{url}
\usepackage{verbatim}
\usepackage{graphicx}
\usepackage{cite}
\usepackage{tabularx}
\usepackage{adjustbox}
\usepackage{booktabs}
\usepackage{multirow}
\usepackage{svg}
\usepackage{comment}
\newcommand{\best}[1]{\textbf{#1}}
\newcommand{\second}[1]{\underline{#1}}
\definecolor{oursblue}{RGB}{220,234,247}
\usepackage{colortbl}
\usepackage[colorlinks=true, urlcolor=black, linkcolor=black]{hyperref}
\newcommand{\red}[1]{{\color{red}#1}}

\title{
SA4Depth: Consistent Pose-Depth Scale Alignment\\for Self-Supervised Monocular Depth Estimation
}

\author{Changxuan Li$^{1,2,3}$, Nadine Berner$^{2}$, Nassir Navab$^{1,3}$, Federico Tombari$^{1,3,4}$, and Stefano Gasperini$^{1,3,5}$%
\thanks{$^{1}$ Technical University of Munich. (email: {\tt\footnotesize firstname.lastname@tum.de})}%
\thanks{$^{2}$ BMW Group. (email: {\tt\footnotesize nadine.berner@bmw.de})}%
\thanks{$^{3}$ Munich Center for Machine Learning (MCML).}%
\thanks{$^{4}$ Google.}%
\thanks{$^{5}$ VisualAIs Labs GmbH.}%
}

\begin{document}


\maketitle
\begin{abstract}
    Self-supervised depth estimation from monocular sequences relies on the joint learning of a depth and a pose network. Despite abundant research done to improve the depth network, efforts on the pose remain limited. In this context, even when depth is estimated up to scale, we highlight the importance of the alignment between the scene scales estimated by the pose and depth nets. Then, we introduce SA4Depth, an approach to improve this alignment and boost the depth predictions while keeping the inference time unchanged. Our proposed method uses the depth estimated during training to reproject learnable visual features across consecutive frames and refine the pose estimates by reducing feature alignment residuals. With our method, the estimated scene scales by the separate depth and pose networks are aligned, and the prediction scale consistency is improved across different sequences. Our differentiable refinement integrates seamlessly into existing self-supervised pipelines and substantially improves their depth estimates. We demonstrate this with extensive experiments both outdoors and indoors on KITTI, Cityscapes, and NYUv2. Additionally, results on KITTI Odometry confirm the effectiveness of our pose refinement. 
    Our code is available at \href{https://github.com/Runningchauncey/SA4Depth}{github.com/Runningchauncey/SA4Depth}.
    
\end{abstract}

\section{Introduction}
\label{sec:intro}

Depth estimation is the foundation of many essential problems in computer vision, such as scene understanding~\cite{zhang2021holistic} and 3D reconstruction~\cite{xu2024depthsplat}. Understanding object distance is crucial for systems interacting with the environment, e.g., in autonomous driving~\cite{Xiang2022-bo} and mobile robotics~\cite{mur2017orb}. Although supervised depth estimation methods have achieved strong results~\cite{bhat2021adabins, yang2024depth}, they learn from ground truth data, which is expensive and time-consuming due to expensive 3D sensors (e.g., LiDAR) and substantial post-processing~\cite{guizilini2020packnet}.

To address these challenges, self-supervised depth estimation methods have leveraged geometric constraints from stereo or monocular video sequences \cite{godard2019digging, zhou2017unsupervised}. Among these, monocular approaches are the most cost-effective~\cite{godard2019digging}, as they require minimal assumptions about the sensor setup, relying solely on image sequences captured by a single camera.

\begin{figure}[t]
    \centering
    \includegraphics[width=0.9\linewidth]{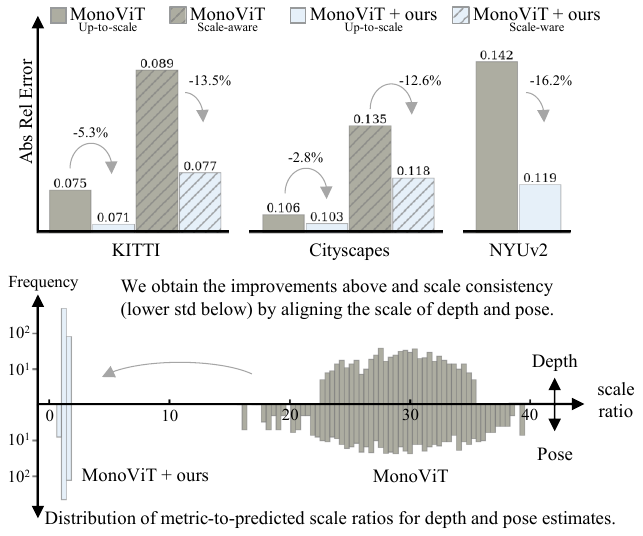}
    \vspace{-1em}
    \caption{We propose SA4Depth, a self-supervised method to make the scale estimation of depth and pose networks aligned and consistent across frames (bottom). Our SA4Depth significantly improves the monocular depth estimation both outdoors~\cite{Geiger2013IJRR,Cordts2016Cityscapes} and indoors~\cite{Silberman:ECCV12} (top). Applicable and effective on many pipelines, the results shown here are on MonoViT~\cite{zhao2022monovit}.}
    \label{fig:teaser}
\vspace{-1.5em}
\end{figure}
Self-supervised monocular methods learn by assuming photometric consistency across consecutive frames~\cite{zhou2017unsupervised}. Extensive research has focused on improving the depth model by introducing new architectures~\cite{zhao2022monovit} and losses~\cite{godard2019digging}.
However, despite its effectiveness in ideal settings, this framework has inherent issues with scale~\cite{guizilini2020packnet,Kinoshita2023-hi}, occlusions~\cite{godard2019digging}, dynamic scenes~\cite{gasperini2021r4dyn}, and adverse weather~\cite{gasperini2023robust}, which prior works have addressed.
Alongside the depth predictions, these approaches predict relative camera poses of frame pairs. Specifically, pose estimates are needed to establish the geometric relationship between adjacent frames, enabling depth learning through photometric consistency without ground truth supervision. Although the pose and depth networks are trained jointly, only the depth net performs inference.

The effectiveness of these methods depends critically on having pose and depth estimates on the same scale \cite{wang2018learning, zhao2020towards}. To learn depth at best, depth and pose need to share a consistent scale so that the photometric loss reflects actual depth errors and not scaling issues. However, Fig.~\ref{fig:teaser} and our analysis in Section~\ref{sec:analysis_scale} show that direct pose regression and inherent scale ambiguity cause large fluctuations across scenes in both depth and pose scales, destabilizing the delicate self-supervision and ultimately degrading the depth estimation performance.
Some works \cite{wang2018learning, zhao2020towards} replace the pose network with a differentiable pose solver, but make it computationally expensive, and are incompatible with common self-supervised frameworks~\cite{godard2019digging,zhao2022monovit}.
A handful of studies improve the pose network itself \cite{Bangunharcana2023-iw}, yet require extra frames at test time.

In this paper, we tackle this problem by improving the pose estimates within the joint depth and pose self-supervised framework, while leaving inference completely unchanged. Toward this end, we show the negative impact of misaligned depth and pose scales within the framework and the benefits of consistent scene scale estimates for better depth outputs. We propose to leverage direct feature alignment to refine the pose estimates. So, we jointly train a feature net to learn robust features for pose refinement and infer a confidence map to mitigate the impact of textureless regions.
We use the estimated depth during training for feature projection, with the feature alignment residuals indicating the misalignment between pose and depth. By reducing these residuals using the Levenberg-Marquardt algorithm, we refine the pose toward higher accuracy and the depth scale. We then use the refined pose to calculate the photometric loss and improve the pose net estimates. By doing so, our method consistently and significantly improves the depth accuracy over many baselines and datasets, and reaches substantial improvements on visual odometry on the KITTI odometry benchmark~\cite{Geiger2012CVPR}. 
Our main contributions can be summarized as follows:
\begin{itemize}
    \item We revisit the role of pose estimation in self-supervised depth training, considering the impact of the alignment between depth and pose in terms of the estimated scene scale, highlighting the importance of aligned pose-depth scale across different sequences.

    \item We propose to couple depth and pose estimation through a pose refinement routine, where we refine the relative pose over the reprojected learnable correspondence feature distance, directly impacting the scale consistency.

    \item Extensive experiments on KITTI~\cite{Geiger2013IJRR}, Cityscapes~\cite{Cordts2016Cityscapes}, and NYUv2~\cite{Silberman:ECCV12} show the effectiveness of our method as it consistently and substantially outperforms the baseline methods both outdoors and indoors.
\end{itemize}

\section{Related Works}
\label{sec:related_works}

\subsection{Self-Supervised Monocular Depth Estimation}

The SSMDE framework was initially proposed by \cite{zhou2017unsupervised}, where the ego-motion and depth are jointly learned and supervised by the photometric loss on the warped view. Monodepth2~\cite{godard2019digging} further improves the loss robustness against occlusions and stationary pixels. Furthermore, feature warping is explored by \cite{zhan2018unsupervised} to enhance the loss robustness and generalization. Extensive research improves the depth accuracy~\cite{Liu2024-wc, yan2021channel, zhao2022monovit} from data or architecture perspectives to reach the performance level of the supervised counterpart. 

While standard approaches learn depth up to scale~\cite {zhou2017unsupervised}, metric scale with self-supervision has been achieved by using additional ubiquitous supervisory signals, such as velocity~\cite{guizilini2020packnet}, camera height~\cite{Xiang2022-bo}, and shape priors of vehicles~\cite{Kinoshita2023-hi}. 


The proxy task of pose estimation within self-supervised depth remains relatively underexplored, yet it is a crucial part of the framework. Commonly, this is conducted by a direct pose regressor~\cite{zhou2017unsupervised}. However, with the limited geometry comprehension and inherent pose ambiguity from 2D images, this strategy yields suboptimal accuracy and poor scale consistency~\cite{wang2018learning}. 
To tackle the pose inaccuracy and scale inconsistency issue, Zhao et al.~\cite{zhao2020towards} replace the pose net with flow predictions, estimating pose from dense correspondences. Indoor depth methods \cite{li2022monoindoor++} chain multiple pose regressors to reduce the pose residuals. However, the relative pose is still inferred directly from 2D image pairs but lacks support from geometry comprehension. Wang et al.~\cite{wang2018learning} leverage differentiable direct visual odometry, but the photometric error used for refinement is not robust, leading to suboptimal performance. A closely related approach is DualRefine~\cite{Bangunharcana2023-iw}, which alternately refines depth and pose via feature matching following epipolar constraints but requires more frames at test time. Our single-frame method does not affect the inference time and closely couples depth and pose estimates through pose refinement via a simple optimization process that better exploits the learned scene geometry.

\subsection{Relative Pose Estimation}
Relative pose estimation aims to infer the 6 Degrees of Freedom transformation from an image pair and is conventionally accomplished by 2D feature matching.
Feature matching predicts keypoints and computes correspondences across the learned features~\cite{sarlin20superglue}. LM-Reloc~\cite{von2020lm} leverages dense feature direct alignment with geometric priors and refines the directly regressed pose. PixLoc~\cite{Sarlin2021-ly} improves the feature direct alignment for large baselines by optimizing across feature scales. We draw inspiration from direct feature alignment methods but drop the requirement for ground truth scene geometry or correspondences to jointly learn depth and pose in an unsupervised manner. 
MADPose~\cite{Yu2025-zo} explores leveraging off-the-shelf 3D geometry estimates for accurate relative pose estimation with sparse correspondences. Instead, we jointly learns the depth and pose toward mutual improvements and utilize dense features for relative pose refinement.

\section{Preliminaries}
\label{sec:preliminary}

The SSMDE framework in \cite{godard2019digging} is revisited in this section. During training, given a triplet of monocular images, for the center frame $\mathcal{I}_{t}$, depth net $f_{\theta_D}$ and pose net $f_{\theta_P}$ separately estimate depth map $D_{t}$ and relative camera poses $P_{t \rightarrow t'}$ to adjacent frames noted as $\mathcal{I}_{t'}$, as:
\begin{equation}
    D_{t}= f_{\theta_D}(\mathcal{I}_{t}),\label{eq:depth_net}
\end{equation}
\begin{equation}
    P_{t \rightarrow t'}= f_{\theta_P}( cat \left[ \mathcal{I}_{t}, \mathcal{I}
    _{t'}\right]), \label{eq:pose_net}
\end{equation}
where $cat \left[\cdot, \cdot\right]$ is channel-wise concatenation. For simplicity, the relative pose will be denoted $P$.

With the depth $D_{t}$ and the camera pose $P$, 
the adjacent frame $\mathcal{I}_{t'}$ can be warped to synthesize $\mathcal{I}_{t}$ by reconstructing the point cloud with known camera intrinsics $K$ and reprojecting to the adjacent frame as
\begin{equation}
    \mathcal{I}_{t' \rightarrow t}= \mathcal{I}_{t'}\langle reproj(recon(D_{t}, K
    ), P) \rangle , \label{eq:warping}
\end{equation}
where $\langle \cdot\rangle$ denotes the bilinear sampling and returns a warped image. 
The photometric loss is calculated as
\begin{equation}
    \mathcal{L}_{ph}= \min_{t^* \in \{t-1, t, t+1\}}pe(\mathcal{I}_{t^* \rightarrow
    t}, \mathcal{I}_{t}). \label{eq:photometric_loss}
\end{equation}
The pairwise photometric error $pe$ is formulated as a weighted sum of the L1 and SSIM loss,
\begin{equation}
    pe(\mathcal{I}_{a, b})= \frac{\alpha}{2}(1 - \text{SSIM}(\mathcal{I}_{a, b})) + (1 - \alpha)\left| \mathcal{I}_{a}- \mathcal{I}_{b}\right
    |.
\end{equation}
An edge-aware smoothness loss is applied for gradient consistency between the depth map and image:
\begin{equation}
    \mathcal{L}_{s}= \left| \partial_{x}d_{t}^{*}\right| e^{-\left|\partial_x \mathcal{I}_t\right|}
    + \left| \partial_{y}d_{t}^{*}\right| e^{-\left|\partial_y \mathcal{I}_t\right|},
    \label{eq:smoothness_loss}
\end{equation}
where $d_{t}^{*}= d_{t}/mean(d_t)$ is mean-normalized disparity $d_{t}$, and the disparity is inverse depth $d_{t}=1/D_{t}$.

\begin{table}[t]
\centering
\caption{Analysis on the impact of pose-depth scale alignment, building upon Md2-50~\cite{godard2019digging} on VKITTI2~\cite{cabon2020virtual}. The 2$^{nd}$ column indicates how the models learn the pose. Scaling factor $s_{depth}: med(D_{gt})/med(D_{pred})$, $s_{pose}: |\tau_{gt}|/|\tau_{pred}|$. \textbf{Best}.}
\vspace{-0.5em}
\label{tab:pose_accuracy_impact}
\resizebox{\linewidth}{!}{%
\begin{tabular}{ll|cccc}
    \toprule 
    No. & Pose in training & Abs Rel & $\delta < 1.25$ & $std(s_{depth})$ & $std(s_{pose})$\\
    \midrule 
    (1)  & self-supervised    & 0.165 & 0.808 & 4.33 & 17.00\\
    (2)  & scaled GT pose    & 0.231 & 0.663 & 0.11 & GT  \\
    (3)  & direct GT pose    & 0.158 & 0.809 & \textbf{0.05} & GT  \\
    (4)  & pose supervised   & 0.153 & \textbf{0.823} & \textbf{0.05} & \textbf{0.30} \\
    (5)  & vel. supervised    & \textbf{0.146} & 0.811 &  \textbf{0.05} & \textbf{0.30} \\
    \bottomrule
\end{tabular}
}
\vspace{-2em}
\end{table}

\section{Analysis on the Scale Alignment}
\label{sec:analysis_scale}
The aligned pose-depth scale is crucial for image warping (Eq.~\ref{eq:warping}), which serves as the base for joint depth and pose learning.
However, due to the monocular setup, both depth and pose estimates are only up to scale. When depth estimates are accurate, pose estimates of a different scale cause erroneous pixel projections and large photometric errors (Eq.~\ref{eq:photometric_loss}), resulting in incorrect gradients for the depth network. Moreover, the scale inconsistency causes depth and pose estimates to change significantly across samples, resulting in ineffective learning and suboptimal performance.

To examine the impact of aligned pose-depth scale, Table~\ref{tab:pose_accuracy_impact} reports depth metrics alongside scale variances. The analysis is performed using the synthetic VKITTI2 dataset~\cite{cabon2020virtual}, which has perfect ground truth (GT). 
\begin{itemize}
    \item Using the GT pose scaled by median scaling factor $s$ of depth estimates (2) skips joint learning and refers to GT depth to align pose scale to current depth estimates. Due to noisy depth predictions at the early training stage, median scaling alone does not ensure accurate scale alignment, underscoring the need for jointly learned pose and a learnable scale alignment.
    \item Using the GT pose directly (3) also skips joint learning but provides a fixed scale, achieving better depth accuracy than (1, 2). But the fixed pose cannot adapt to depth estimates as in joint learning, the depth network must learn both depth details and metric scale from the photometric loss, resulting in worse performance than (4, 5). 
    \item Supervising the pose with the GT pose (4) or GT translation norm (5) as in~\cite{guizilini2020packnet} achieves comparable depth error reduction and low variance in both depth and pose scale factors. In these cases, the pose, unlike in (3), can adapt to the learned depth in joint learning and is constrained to stabilize the estimation scale.
\end{itemize}

These experiments indicate that jointly learning depth and pose with constraints on the pose scale is beneficial, and that explicit scale constraints improve scale consistency.

\begin{figure*}[ht]
    \centering
    \includegraphics[width=1.0\textwidth]{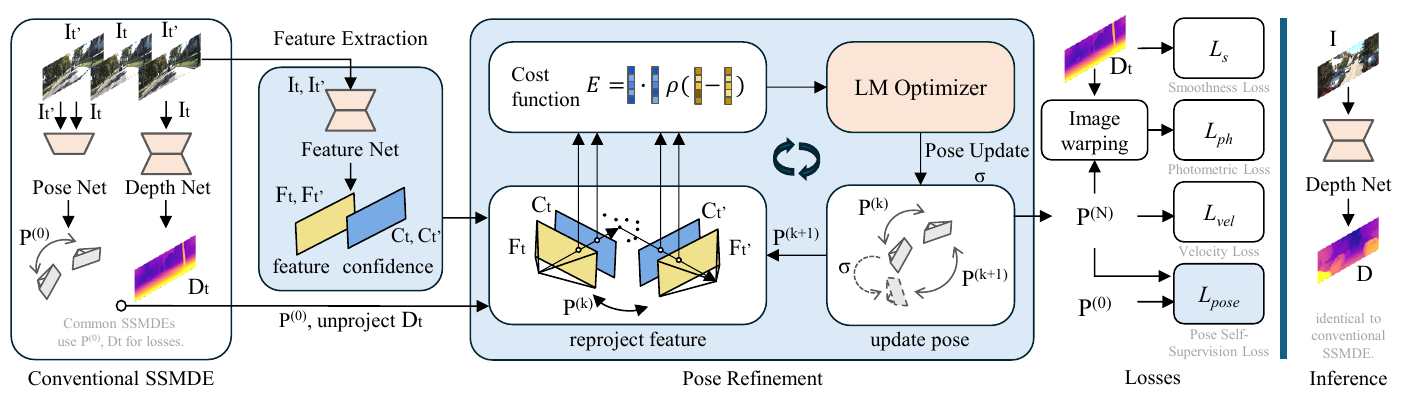}
    \vspace*{-2em}
    \caption{Overview of the proposed SA4Depth, with our contributions to SSMDE highlighted in light blue. Compared to conventional SSMDE(left), our approach refines the initial pose $P^{(0)}$ using the estimated depth $D_{t}$, learned feature $F_{\{t,t'\}}$, and confidence $C_{\{t,t'\}}$ maps.
    Our training losses use the depth $D_{t}$ and the refined pose $P^{(N)}$. The inference is unchanged (right).
    At training time, our method takes an image triplet as input, with the center frame $\mathcal{I}_{t}$ and the adjacent frames $\mathcal{I}_{t'}$. A depth net estimates depth for $\mathcal{I}_{t}$ and a pose net estimates the camera pose of $\mathcal{I}_{t}$, referring to each $\mathcal{I}_{t'}$. A feature net extracts a feature and confidence map for each image. Our pose refinement reprojects $F_{t}$ by depth map into an adjacent frame and aligns it with the adjacent frame feature $F_{t'}$ by refining from $P^{(0)}$ to $P^{(N)}$. It iteratively updates the pose by optimizing the cost function $E$ using the Levenberg-Marquardt (LM) algorithm. Additionally, we supervise the pose net using the refined pose with $\mathcal{L}_{pose}$ and optionally $\mathcal{L}_{vel}$ to learn metric depth.}
    \label{fig:overview}
    \vspace*{-1em}
\end{figure*}

\section{Method}
\label{sec:method}
\textbf{Overview of SA4Depth}
Our method builds upon the conventional SSMDE pipeline~\cite{zhou2017unsupervised,godard2019digging}.
The general pipeline is illustrated in Fig.~\ref{fig:overview}. In particular, our pose refinement is similar to that of PixLoc~\cite{Sarlin2021-ly}. Our contributions lie in integrating it into the SSMDE framework by using the current depth estimates for feature reprojection, taking the current pose net output as initialization, and self-supervising the pose net with the refined pose. 




\textbf{Feature extraction}
\label{sec:method_refinement}
We utilize deep visual features as the metric for the reprojection error caused by misaligned depth and pose (as opposed to the sensitive photometric error in \cite{wang2018learning}) and employ this reprojection error to refine the pose with knowledge from the current depth prediction.

So, we extract a feature map $F$ and a confidence map $C$ from each image $\mathcal{I}$ using a feature net $\theta_{F}$:
\begin{equation}
    F, C= {\theta_{F}}(\mathcal{I}). \label{eq:feature_net}
\end{equation}

We use the confidence maps $C$ to downweight dynamic objects or textureless areas that disrupt the reprojection. The feature net is built upon a VGG-19 backbone.

\textbf{Pose refinement}
We formulate the pose refinement as an optimization problem with a cost function $E$ based on the weighted feature distance between reprojected point pairs:
\begin{equation}
    E(P)= \frac{1}{\mathcal{M}}\sum c_{ref}\cdot
    c_{query}\cdot \rho (\left| \delta\right|^{2}_{2}), 
\end{equation}
where $\delta = f_{query} - f_{ref}$ is the feature difference between the query feature $f_{query}$ (end of reprojection) and the reference feature $f_{ref}$ (start of reprojection). The total cost for an image pair is computed as the robust loss $\rho$ over all reprojection pairs, weighted by the confidence values $c_{query}$ and $c_{ref}$. For efficiency, we sample only $\mathcal{M}$ points in the reference frame $\mathcal{I}_{ref}$ to compute the cost $E$. We take the middle frame $\mathcal{I}_{t}$ as a reference frame where the depth map is estimated, and adjacent frames $\mathcal{I}_{t'}$ as query frames.

The relative pose variation changes the coordinate of a 3D point $s_{query}$ in the query frame, therefore shifting its projected location $i_{query}$ on the query image plane. Consequently, a different feature vector $f_{query}$ is interpolated from the query frame's feature map $F_{query}$. The Jacobian of the feature difference $\delta$ with respect to the pose update $\sigma$ is thus obtained via the chain rule~\cite{Sarlin2021-ly}:
\begin{equation}
    J = \frac{\partial F_{query}}{\partial \sigma}= \frac{\partial F_{query}}{\partial
    i_{query}}\cdot\frac{\partial i_{query}}{\partial s_{query}}\cdot\frac{\partial s_{query}}{\partial
    \sigma},
\end{equation}
where the Jacobian is the chaining feature gradient, the projection function derivatives, and the coordinate transformation Jacobian. Pose updates are solved in closed form via Levenberg-Marquardt optimization:
\begin{equation}
    \sigma = -(\mathcal{H}+ \lambda \cdot diag(\mathcal{H}))^{-1}J^{T}\mathcal{W}\Delta
    , \label{eq:pose_update}
\end{equation}
where the Hessian $\mathcal{H} = J^{T} \mathcal{W} J$ is weighted by $\mathcal{W}$, $\lambda$ is a learnable damping factor adjusting to ego-motion distribution across the dataset, and $\Delta$ stacks all feature differences $\delta$. The diagonal weighting matrix $\mathcal{W}$ is defined as:
\begin{equation}
    \mathcal{W}= to\_diag(\left[c_{ref}\cdot
    c_{query}\cdot \rho '(\left| \delta\right |^{2}_{2})\right ]_{\mathcal{M}} ),
\end{equation}
where $\rho'$ denotes the robust loss function derivatives, and confidence scores weigh feature differences.

To refine confidence weighting during iterations, we adopt an Iteratively Reweighted Least Squares (IRLS) to dynamically reweight the samples in iterations as follows:
\begin{equation}
    \mathcal{W}^{(k+1)}= \mathcal{W}^{(k)} \cdot \frac{\rho (\left| \Delta\right|^{2}_{2})}{\rho (\left| \Delta^{(k)} + J \sigma^{(k)} \right|^{2}_{2})}
    \label{eq:IRLS}
\end{equation}
where the weight matrix for the next iteration $k+1$ is updated by the ratio of the previous cost over the updated cost by the new iteration step $\sigma^{(k)}$. This ensures a larger weight is assigned to more influential point positions.

The pose is then updated via exponential mapping:
\begin{equation}
    P^{(k+1)}= exp(\sigma^{(k)\wedge}) P^{(k)},
\end{equation}
where $\wedge$ represents the skew-symmetric matrix, and $exp$ denotes mapping to $SE(3)$. This iterative refinement process continues for $N$ iterations.
The pose refinement can be viewed as an internalized SLAM. The pose optimization in SLAM methods~\cite{tateno2017cnn} is typically performed at test time. We apply our pose refinement during training, achieving consistent scale alignment without increasing the runtime.

\textbf{Coupling between depth and pose}
The depth estimates define the 3D point cloud for feature reprojection, which ultimately influences the pose updates.
Therefore, similar to \cite{wang2018learning}, when backpropagating the photometric loss $\mathcal{L}_{ph}$, the depth estimates $D$ receive gradients $\nabla_{D} \mathcal{L}_{ph}$ both directly from the photometric loss and via the camera pose $P$, as:
\begin{equation}
    \nabla_{D} \mathcal{L}_{ph} = \frac{\partial \mathcal{L}_{ph}}{\partial D} + \frac{\partial \mathcal{L}_{ph}}{\partial P} \frac{\partial P}{\partial D}.
    \label{eq:depth_grad}
\end{equation}

This gradient formulation creates a coupling between depth and pose. The pose scale is constrained by the online depth estimates and refined to align with the depth scale. Meanwhile, the depth scale converges and remains stable through the indirect gradient via pose.

\begin{table*}
[ht]
\caption{\label{tab:kitti_improved}Comparison on the KITTI~\cite{Geiger2013IJRR} Eigen split with improved GT~\cite{uhrig2017sparsity}. Resolution of 640 $\times$ 192. Upper half for results median-scaled by depth GT. Lower half for learned metric depth. Training: \textbf{M} monocular videos, \textbf{VI} video interpolation, \textbf{v} w/ velocity, \textbf{SI} scraping dataset, \textbf{camH} w/ camera height. All results are \textbf{w/o post-processing}. We reproduced \cite{godard2019digging,zhao2022monovit}. 18 and 50 are ResNet~\cite{he2016deep} sizes. Scale Std: $std$ of depth scaling factor. Highlighting: \best{Best}, \second{Second}.}
\vspace{-0.7em}
\centering
\resizebox{\textwidth}{!}{%
\begin{tabular}{l|ll|c|cccccccccccc}
\toprule
Scaling & Method & Training  & Scale Std & Abs Rel & Sq Rel & RMSE & RMSE log & $\delta < 1.25$ & $\delta < 1.25^{2}$ & $\delta < 1.25^{3}$ \\
\midrule
\multirow{11.5}{*}{\shortstack{done \\w/ GT}} & PackNet-SfM~\cite{guizilini2020packnet} & M    & $0.75$ & 0.078                & 0.420               & 3.485             & 0.121                 & 0.931                      & 0.986                          & 0.996                          \\
                              & DualRefine~\cite{Bangunharcana2023-iw} & M & $3.39$ & 0.075 & 0.379 & 3.490 & 0.117 & 0.936 & 0.989 & \second{0.997} \\
                              & Mono-ViFI~\cite{Liu2024-wc}    & M+VI  &$3.32$   & 0.072       & 0.342        & \second{3.254}      & 0.111          & 0.941   & \second{0.991}                   & \best{0.998}                   \\
\cmidrule{2-11}               
                              & Md2-18~\cite{godard2019digging} & M   & $2.79$   & 0.090                                 & 0.545 & 3.942 & 0.137                 & 0.914                      & 0.983                          & 0.995                          \\
                              & \cellcolor{oursblue}\textbf{+ ours} & 
                              \cellcolor{oursblue}M   & 
                              \cellcolor{oursblue}$\best{0.09}$   & 
                              \cellcolor{oursblue}0.087                & 
                              \cellcolor{oursblue}0.524               & 
                              \cellcolor{oursblue}3.910             & 
                              \cellcolor{oursblue}0.134                 & 
                              \cellcolor{oursblue}0.918                      & 
                              \cellcolor{oursblue}0.982                          & 
                              \cellcolor{oursblue}0.995                          \\
\cmidrule{2-11}
                              & Md2-50~\cite{godard2019digging} & M   & $2.65$   &0.085  &   0.468  &   3.674  &   0.128  &   0.921  &   0.985  &   0.996  \\
                              & \cellcolor{oursblue}\textbf{+ ours} &
                              \cellcolor{oursblue}M & 
                              \cellcolor{oursblue}$1.90$ & 
                              \cellcolor{oursblue}0.080     & 
                              \cellcolor{oursblue}0.465     & 
                              \cellcolor{oursblue}3.707     & 
                              \cellcolor{oursblue}0.125     & 
                              \cellcolor{oursblue}0.928           & 
                              \cellcolor{oursblue}0.986               & 
                              \cellcolor{oursblue}0.996 \\
\cmidrule{2-11}
                              & MonoViT~\cite{zhao2022monovit} &
                              M   & 
                              $ 3.73$ &
                              0.075 &
                              0.389  &
                              3.419  &
                              0.115 &
                              0.938 &
                              0.989 &
                              \second{0.997}                 \\
                              & \cellcolor{oursblue}\textbf{+ ours}    &
                              \cellcolor{oursblue}M  & 
                              \cellcolor{oursblue}$ \second{0.16}$    & 
                              \cellcolor{oursblue}\second{0.071}         & 
                              \cellcolor{oursblue}0.380                &
                              \cellcolor{oursblue}3.341   & 
                              \cellcolor{oursblue}0.111          & 
                              \cellcolor{oursblue}\second{0.945}     & 
                              \cellcolor{oursblue}0.989                 & 
                              \cellcolor{oursblue}\second{0.997}                          \\
\cmidrule{2-11}
                              & Hybrid-depth~\cite{zhang2025hybrid}&
                              M   & 
                              3.56 &   
                              0.072  &   
                              \second{0.334}  &   
                              3.264  &   
                              \second{0.110}  &   
                              0.944  &   
                              \second{0.991}  &   
                              \textbf{0.998}  \\     
                              & \cellcolor{oursblue}\textbf{+ ours} &
                              \cellcolor{oursblue}M  & 
                              \cellcolor{oursblue}2.71     & 
                              \cellcolor{oursblue}\textbf{0.068}  &   
                              \cellcolor{oursblue}\textbf{0.317}  &   
                              \cellcolor{oursblue}\textbf{3.103}  &   
                              \cellcolor{oursblue}\textbf{0.105}  &   
                              \cellcolor{oursblue}\textbf{0.952}  &   
                              \cellcolor{oursblue}\textbf{0.992}  &   
                              \cellcolor{oursblue}\textbf{0.998}  \\
\midrule
\multirow{9}{*}{\shortstack{scale-\\aware}}  & VADepth~\cite{Xiang2022-bo}& M+camH & $0.07$ & 0.093                & 0.493               & 3.704             & 0.132                 & 0.913                      & 0.984                          & \second{0.996}                 \\
                              & FUMET~\cite{Kinoshita2023-hi}  & M+SI  & $0.07$ & 0.091                & 0.533               & 3.892             & 0.136                 & 0.906                      & 0.983                          & 0.995                          \\
\cmidrule{2-11}                
                              & Md2-18~\cite{godard2019digging}  & M+v & $0.07$  & 0.092 & 0.549 & 3.995 & 0.142 & 0.902 & 0.980 & 0.994                          \\
                              & \cellcolor{oursblue}\textbf{+ ours}  &
                              \cellcolor{oursblue}M+v  & \cellcolor{oursblue}$\second{0.06}$  & \cellcolor{oursblue}0.088 & \cellcolor{oursblue}0.556 & \cellcolor{oursblue}3.938 & \cellcolor{oursblue}0.138 & \cellcolor{oursblue}0.910 & \cellcolor{oursblue}0.980 & \cellcolor{oursblue}0.994                         \\
\cmidrule{2-11}
                              & Md2-50~\cite{godard2019digging}  & M+v &  $0.07$  & 0.093                & 0.559               & 3.984             & 0.140                 & 0.901                      & 0.981                          & 0.995                          \\
                              & \cellcolor{oursblue}\textbf{+ ours} &
                              \cellcolor{oursblue}M+v  & \cellcolor{oursblue}$\second{0.06}$ & \cellcolor{oursblue}\second{0.085} & \cellcolor{oursblue}0.476 & \cellcolor{oursblue}3.782 & \cellcolor{oursblue}0.133 & \cellcolor{oursblue}\second{0.914} & \cellcolor{oursblue}0.983 & \cellcolor{oursblue}0.995                          \\
\cmidrule{2-11}
                              & MonoViT~\cite{zhao2022monovit} & M+v &  $0.07$   & 0.089                & \second{0.466}      & \second{3.590}    & \second{0.129}        & 0.913                      & \second{0.985}                 & \second{0.996}                 \\
                              & \cellcolor{oursblue}\textbf{+ ours}    &
                              \cellcolor{oursblue}M+v & \cellcolor{oursblue}$\best{0.05}$    & \cellcolor{oursblue}\best{0.077}         & \cellcolor{oursblue}\best{0.404}        & \cellcolor{oursblue}\best{3.472}      & \cellcolor{oursblue}\best{0.119}          & \cellcolor{oursblue}\best{0.932}               & \cellcolor{oursblue}\best{0.988}                   & \cellcolor{oursblue}\best{0.997}                   \\
\bottomrule
\end{tabular}
}
\vspace*{-1.5em}
\end{table*}

\textbf{Loss functions}
\label{sec:loss}
We make our pose refinement differentiable to keep the whole pipeline trainable end-to-end. 
During training, as the pose estimate improves with our refinement steps, we use this refined pose for image warping, as shown in Fig.~\ref{fig:overview}.
Then, as we have a better pose estimate $P_{t \rightarrow t'}^{(N)}$ compared to the initial estimate $P_{t \rightarrow t'}^{(0)}$ from the pose net, we use the refined one to self-supervise the pose net. Toward this end, we define the following pose self-supervision loss:
\begin{equation}
    \mathcal{L}_{pose}(P^{(0), (N)}) = \left|\tau^{(0)}-\tau^{(N)}\right|+\left|R^{(0)}-R^{(N)}\right|,
    \label{eq:pose_sup}
\end{equation}
where $\tau$ and $R$ are the translation vector and rotation matrix of the pose of the corresponding superscript.

Furthermore, we optionally use velocity supervision from \cite{guizilini2020packnet} for scale-awareness, which acts on the magnitude of the translation. In particular, we apply it to our refined pose $P^{(N)}$ instead of on the pose net output. The velocity loss is as 
\begin{equation}
    \mathcal{L}_{v}(\tau^{(N)}, v) = \begin{vmatrix} \| \tau^{(N)}\|_{2} - \left| v \right| dt \end{vmatrix},
    \label{eqn:velocity_loss}
\end{equation}
where the predicted translation norm $\| \tau^{(N)}\|_{2}$ is compared to the ground truth translation norm by the ego vehicle velocity $v$ and the time interval $dt = |t - t'|$.

The total training loss is a weighted sum of the photometric loss $\mathcal{L}_{ph}$, smoothness loss $\mathcal{L}_{s}$, pose self-supervision loss $\mathcal{L}_{pose}$, and optionally velocity loss $\mathcal{L}_{v}$, as
\begin{equation}
    \mathcal{L}= \mathcal{L}_{ph}+ \beta_{s}\mathcal{L}_{s}+\mathcal{L}_{pose}+\beta_{v}\mathcal{L}_{v},
\end{equation}
where $\beta_{s}=1e-3$ following \cite{godard2019digging}, and $\beta_{v}=0.0$ for up-to-scale depth or $\beta_{v}=0.02$ for metric depth as in~\cite{gasperini2023robust}.

\textbf{Implementation Details}
\label{sec:implementation}
We implement our model in PyTorch, trained for
20 epochs using Adam~\cite{kingma2014adam} with a batch size of 12. The learning rate is $10^{-4}$ and drops to $10^{-5}$
after 15 epochs.
With high compatibility, we apply our method on top of the baselines, inheriting their depth and pose net designs, where the rotation is represented with an axis-angle.
\section{Experiments and Results}
\label{sec:experiments}
\subsection{Experimental Setup}
\label{sec:setup}

\textbf{Datasets and metrics}
We evaluated our methods on KITTI~\cite{Geiger2013IJRR}, Cityscapes~\cite{Cordts2016Cityscapes}, and NYUv2~\cite{Silberman:ECCV12}. For the driving datasets KITTI and Cityscapes, we train our models in both up-to-scale and scale-aware setups by switching off/on the velocity loss of Eq.~\ref{eqn:velocity_loss}, which uses the widely available and easy to acquire ego vehicle velocity signal.
For \textbf{KITTI} 
, we follow the Eigen train-val split~\cite{eigen2015predicting} and pre-processing in \cite{godard2019digging}, resulting in 39,810 monocular triplets for training and 4,424 for validation. For evaluation, we follow the Eigen split for 652 images with improved ground truth depth by \cite{uhrig2017sparsity} in Tab.~\ref{tab:kitti_improved}.
\textbf{Cityscapes} is a driving dataset of challenging dynamic scenes. We follow \cite{watson2021temporal} to process 69,731 training triplets and 1,525 images for evaluation. 
\textbf{NYUv2} is a challenging indoor dataset recorded with hand-held RGB-D cameras. We follow \cite{li2022monoindoor++} for preprocessing and evaluation on 654 annotated test samples.
We also evaluate our method on \textbf{KITTI Odometry}~\cite{Geiger2012CVPR}, following \cite{zhou2017unsupervised} in using sequences 00 to 08 for training, and sequences 09 and 10 for evaluation.


\textbf{Baselines}
Our method is compatible with various self-supervised pipelines. To showcase this flexibility, we apply it on Monodepth2~\cite{godard2019digging},
MonoViT~\cite{zhao2022monovit}, Mono-ViFI~\cite{Liu2024-wc}, and Hybrid-depth~\cite{zhang2025hybrid}, following their training setups.
%
\subsection{Quantitative Results}
\textbf{Depth on KITTI}
In Table~\ref{tab:kitti_improved}, we report the results of monocular depth estimation on KITTI~\cite{Geiger2013IJRR}, comparing with the state-of-the-art SSMDE methods. Notably, our method improves upon the baselines it is applied.
For the up-to-scale setup (top half), the estimates are median-scaled by the depth GT. Here, our SA4Depth paired with Hybrid-depth~\cite{zhang2025hybrid} outperforms the other works on all metrics.
DualRefine~\cite{Bangunharcana2023-iw} refines an extra frame at test time, so we compare with their single-frame estimates, and ours performs better.
Our work systematically improves the scale consistency over all baselines (represented by the standard deviation in the Scale Std column).
The scale-aware setup (lower half) is more challenging as methods need to learn depth and metric scale simultaneously. FUMET~\cite{Kinoshita2023-hi} also uses extra training data from the web for scale awareness. 
Although FUMET performs slightly better than the Md2-50 it builds on (with Md2 using the weak velocity loss), our method outperforms FUMET by an even larger gap without any extra training data. 
Remarkably, paired with MonoViT, our approach achieves the best results on all metrics.
We also evaluate video depth as in VDA [37] and measure scale consistency using the scale standard deviation across the whole sequence; notably, our method paired with MonoViT outperforms the state-of-the-art VDA model by 34\%, achieving exceptional scale consistency even though ours is trained without GT supervision.
Overall, these results demonstrate the positive impact of applying our method on the current self-supervised depth framework and the value of our insights on scale alignment and consistency.

\textbf{Depth on Cityscapes}
In Table~\ref{tab:cityscape_results}, we provide a comparison on Cityscapes. Our models improve all baselines on key metrics. Notably, in the scale-aware setup, we reduce the absolute relative error by 12.6\% over the baseline MonoViT. 

\begin{table}
    [t]
    \caption{Comparison on Cityscapes~\cite{Cordts2016Cityscapes} at 416 $\times$ 128. All results are \textbf{w/o post-processing}. Notations as in Table~\ref{tab:kitti_improved}.}
    \label{tab:cityscape_results}
    \centering
    \resizebox{\columnwidth}{!}{%
    \begin{tabular}{l|l|ccccc}
    \toprule
    Scaling                       & Method                                            & Abs Rel & Sq Rel & RMSE & RMSE log & $\delta < 1.25$ \\
    \midrule
    \multirow{5.5}{*}{\shortstack{done \\w/ GT}} 
        & Md2-18~\cite{godard2019digging}      
        & 0.122 & 1.355 & 6.613 & 0.180 & 0.860 \\
        & \cellcolor{oursblue}Md2-18 \textbf{+ ours}        
        & \cellcolor{oursblue}0.111                 
        & \cellcolor{oursblue}1.243 
        & \cellcolor{oursblue}6.514     
        & \cellcolor{oursblue}0.173
        & \cellcolor{oursblue}0.874  \\
    \cmidrule{2-7}               
        & MonoViT~\cite{zhao2022monovit}         
        & 0.106   & 1.023 & 6.284 & 0.164 & 0.873 \\
        & \cellcolor{oursblue}MonoViT \textbf{+ ours}         
        & \cellcolor{oursblue}\second{0.103}
        & \cellcolor{oursblue}1.179         
        & \cellcolor{oursblue}5.954
        & \cellcolor{oursblue}0.161
        & \cellcolor{oursblue}\textbf{0.894} \\
    \cmidrule{2-7}               
        & Mono-ViFI~\cite{Liu2024-wc}           
        & 0.105 & \second{0.880} & \textbf{5.550} & \second{0.156} & 0.881  \\
        & \cellcolor{oursblue}Mono-ViFI \textbf{+ ours}
        & \cellcolor{oursblue}\textbf{0.102}
        & \cellcolor{oursblue}\textbf{0.825}         
        & \cellcolor{oursblue}\second{5.561} 
        & \cellcolor{oursblue}\textbf{0.152} 
        & \cellcolor{oursblue}\second{0.885} \\
    \midrule
    \multirow{6.2}{*}{\shortstack{scale-\\aware}}   
        & FUMET~\cite{Kinoshita2023-hi}   
        & \second{0.129} &  \second{1.445} & \second{6.573} & \textbf{0.191} & 0.855 \\
        & Md2-18~\cite{godard2019digging}     
        & 0.140 & 1.904 & 6.807 & 0.196 & 0.852 \\
        & \cellcolor{oursblue}Md2-18 \textbf{+ ours}  
        & \cellcolor{oursblue}0.124
        & \cellcolor{oursblue}1.484 
        & \cellcolor{oursblue}6.687 
        & \cellcolor{oursblue}0.192 
        & \cellcolor{oursblue}0.865 \\
        \cmidrule{2-7}
        & MonoViT~\cite{zhao2022monovit}   & 0.135   & 1.756 & 6.697 & 0.201 & 0.857 \\
        & \cellcolor{oursblue}MonoViT \textbf{+ ours}         
        & \cellcolor{oursblue}\best{0.118}       
        & \cellcolor{oursblue}\best{1.296}         
        & \cellcolor{oursblue}\best{6.283} 
        & \cellcolor{oursblue}\textbf{0.191}
        & \cellcolor{oursblue}\best{0.879} \\
    \bottomrule
    \end{tabular}
    }
    \vspace{-1em}
\end{table}



\textbf{Depth on NYUv2 (indoors)}
In Table~\ref{tab:nyu}, we show the indoor results on NYUv2~\cite{Silberman:ECCV12} of our method paired with different baselines. Despite the SSMDE challenges of NYUv2 due to the complex egomotion \cite{bian2021auto}, our method effectively handles them and improves over the baselines by a substantial margin. Paired with MonoViT, our approach reduces the AbsRel error by 16.2\%. These experiments, along with those on KITTI and Cityscapes, demonstrate the effectiveness of our SA4Depth both indoors and outdoors.

\begin{table}[t]
    \caption{\label{tab:nyu} \textbf{Indoor results} on NYUv2~\cite{Silberman:ECCV12}. Resolution of 320$\times$256.}
    \vspace{-0.5em}
    \centering
    \resizebox{0.9\columnwidth}{!}{%
    \begin{tabular}{l|ccccccc}
    \toprule
    Method                                        & Abs Rel & Log10 & RMSE & $\delta < 1.25$ \\
    \midrule
    Md2-18~\cite{godard2019digging} & 0.156 & 0.066 & 0.577 & 0.779 \\
    \cellcolor{oursblue}\textbf{+ ours}                                 & \cellcolor{oursblue}\textbf{0.136} & \cellcolor{oursblue}\textbf{0.058} & \cellcolor{oursblue}\textbf{0.538} & \cellcolor{oursblue}\textbf{0.826} \\
    \midrule
    Md2-50~\cite{godard2019digging} & 0.154  & 0.065 & 0.581 & 0.784 \\
    \cellcolor{oursblue}\textbf{+ ours}                                 & \cellcolor{oursblue}\textbf{0.133} & \cellcolor{oursblue}\textbf{0.056} & \cellcolor{oursblue}\textbf{0.535} & \cellcolor{oursblue}\textbf{0.838} \\
    \midrule
    MonoViT~\cite{zhao2022monovit}                                      & 0.142 & 0.061 & 0.547 & 0.813 \\
    \cellcolor{oursblue}\textbf{+ ours}                                      & \cellcolor{oursblue}\textbf{0.119} & \cellcolor{oursblue}\textbf{0.051} & \cellcolor{oursblue}\textbf{0.493} & \cellcolor{oursblue}\textbf{0.864} \\
    \bottomrule
    \end{tabular}
    }
    \vspace{-2em}
\end{table}

\textbf{Visual odometry on KITTI Odometry}
In Table~\ref{tab:kitti_odom}, we show the comparison on the KITTI Odometry dataset with both methods specialized on visual odometry (upper) and those focusing on SSMDE like ours (lower). The trajectory estimates are aligned to the ground truth by 7DoF alignment. The average errors in translation and rotation are calculated following \cite{zhan2020visual}. The translational error $t_{err}$ of ours reaches the lowest among the SSMDE methods and is on par with visual odometry methods. A good improvement over both metrics is achieved on top of the baseline Md2~\cite{godard2019digging} with ResNet-50.

\begin{table}[t]
    \caption{Comparison on KITTI \textbf{Odometry} sequences 09-10 to visual odometry methods (upper) and SSMDE methods (lower).}
    \label{tab:kitti_odom}
    \vspace{-0.5em}
    \centering
    \resizebox{\columnwidth}{!}{%
    \begin{tabular}{l|cc|cc}
        \toprule
         \multirow{2}{*}{Method} &  \multicolumn{2}{c|}{Sequence 09}  & \multicolumn{2}{c}{Sequence 10} \\
         &  $t_{err}(\%)$  & $r_{err}(\frac{\circ}{100m})$ &  $t_{err}(\%)$  & $r_{err}(\frac{\circ}{100m})$ \\
        \midrule
        ORB-SLAM2~\cite{mur2017orb} & 3.22 & 0.40 & 4.25 & 0.30 \\
        DF-VO~\cite{zhan2020visual} & 2.40 & 0.24 & 1.82 & 0.38 \\
        \midrule
        SC-Depth~\cite{bian2019unsupervised} & 7.31 & 3.05 & 7.79 & 4.90 \\
        Trianflow~\cite{zhao2020towards} & 6.93 & \best{0.44} & \second{4.66} & \best{0.62} \\
        DualRefine~\cite{Bangunharcana2023-iw} & \second{3.43} & 1.04 & 6.80 & \second{1.13} \\
        Md2-50~\cite{godard2019digging} & 5.26 & 2.06 & 7.91 & 3.64 \\
        \cellcolor{oursblue}Md2-50 \textbf{+ ours} & \cellcolor{oursblue}\best{2.96} & \cellcolor{oursblue}\second{0.96} & \cellcolor{oursblue}\best{3.48} & \cellcolor{oursblue}1.91 \\
        \bottomrule
    \end{tabular}
    }
\vspace{-2em}
\end{table}

\textbf{Runtime}
Our work improves pose estimation within a self-supervised learning framework. As the depth model is unchanged, the runtime of depth inference is identical to the baseline. In Table~\ref{tab:ablations} (rightmost), we show the run time on an NVIDIA T4 GPU. Our SA4Depth delivers superior depth estimates without any inference time drawbacks. 


\textbf{Comparison with solver-based pose estimators}
In Table~\ref{tab:ablations} (7-9), we show the impact of replacing the pose net with SuperGlue~\cite{sarlin20superglue} as a feature matcher paired with the pose solver of RANSAC (7) or PnP + RANSAC (8) to estimate the camera pose. Although for the pure pose estimation task, these methods work well, due to the large-scale discrepancy between pose and depth estimates, they harm the depth learning. Similarly to our pose refinement, we use SuperGlue and RANSAC (9) to refine the pose estimates. However, depth is poorly learned in this setup, yielding worse results than the baseline (1) and ours (6).

\begin{table}[b]
    \vspace{-1em}
    \caption{Ablation study using Md2-50~\cite{godard2019digging}. Same setup as in Table~\ref{tab:kitti_improved}, with estimates up-to-scale. FPS indicates inference speed.}
    \label{tab:ablations}
    \vspace{-0.5em}
    \centering
    \setlength{\tabcolsep}{3pt}
    \resizebox{\linewidth}{!}{%
    \begin{tabular}{ll|ccc}
        \toprule No. & Method                                           & Abs Rel & $\delta < 1.25$  & FPS \\
        \midrule
        (1) &  Md2-50~\cite{godard2019digging}                                    & 0.086   & 0.919           & 61        \\
        (2) & pose by PixLoc~\cite{Sarlin2021-ly} w/ LiDAR points                           & 0.398 & 0.365               & 61        \\
        (3) & (2) w/ pred. depth, no LiDAR & 0.086 & 0.920 & 61\\
        (4) & + pose net as init. & 0.083 & 0.923               & 61 \\
        (5) & + pose net self-sup (Eq.~\ref{eq:pose_sup})  & 0.082 & 0.926           & 61  \\
        \cellcolor{oursblue}(6) & \cellcolor{oursblue}+ IRLS (Eq.~\ref{eq:IRLS}) = \textbf{SA4Depth [ours]} & \cellcolor{oursblue}\textbf{0.080} & \cellcolor{oursblue}\textbf{0.928}           & \cellcolor{oursblue}61 \\
        \midrule 
        (7) & pose by SG~\cite{sarlin20superglue} + RANSAC & 0.418 & 0.321 & 61 \\
        (8) & (7) + PnP & 0.418 & 0.321 & 61\\
        (9) & (7) + pose net & 0.239 & 0.596  & 61 \\
        \midrule
        (10) & (6), but detach depth for feat. reproj  & 0.118 & 0.857 & 61\\
        (11) & (6), but refined pose only for $\mathcal{L}_{pose}$ & 0.096 & 0.910 & 61\\
        \midrule
         (12) & (6) w/o feature map    & 0.089 & 0.916 & 61\\
         (13) & (6) w/o confidence map   & \second{0.081} & 0.925 & 61\\
         (14) & (6) w/o learnable $\lambda$ in Eq.10 & 0.082 & \textbf{0.928} & 61\\
         \midrule
         (15) & (6) w/ MASt3R~\cite{leroy2024grounding} feature & 0.082 & \second{0.927} & 61\\
         (16) & (6) w/ PixLoc~\cite{Sarlin2021-ly} feature & 0.087 & 0.920 & 61\\
        \bottomrule
    \end{tabular}
    }
\end{table}

\textbf{Ablation study}
\label{sec:ablations}
In Table~\ref{tab:ablations}, we report an ablation study to show the impact of the main components of our method. Starting from Md2-50 (1), we add a straightforward adaptation of PixLoc~\cite{Sarlin2021-ly}, which replaces the pose net for pose estimation using LiDAR points for image feature reprojection (2). This results in significantly worse depth estimates. This demonstrates how PixLoc is not compatible with such self-supervised frameworks out of the box, but adaptations are necessary. Then, with (3), we use the predicted depth instead of the LiDAR data to reproject the features, but it performs on par with the baseline. Then, we use the current pose estimates as initialization for the refinement process, replacing the coarse-scale pose refinement in PixLoc (4). This brings a small improvement. Furthermore, we self-supervise the pose net with the updated pose estimate from the refinement (5), which further improves the depth. Finally, with (6), we dynamically reweight the feature alignment during the refinement, which enhances the depth estimates even more (our SA4Depth). Overall, Table~\ref{tab:ablations} demonstrates the necessity of our modifications to outperform the baselines, showing that feature matching (7-9) or direct feature alignment (2-3) alone are not sufficient. 

Then, we investigate the influence of the additional gradient on the depth estimator from Eq.~\ref{eq:depth_grad} by detaching the depth map's gradients in the feature reprojection (10) and using the refined pose only for pose self-supervision (11). That means there's no indirect gradient from the pose to the depth, but only a better-aligned pose (10) or extra supervision on the pose (11). These are proven insufficient to improve the depth learning, with the performance degrading due to under-constrained estimated scales.

In addition, we analyze the role of pose refinement components (12-14) and choices of the feature map (15-16). We remove in the pose refinement every time a single element as feature maps (12), confidence maps (13), and learnable damping factors (14). Replacing the learnable feature map with images (12) turns the feature-based pose refinement into photometric pose refinement, resulting in inaccuracy in reflecting the pose-depth misalignment and a large performance drop. Removing the confidence map (13) or learnable damping factor $\lambda$ from Eq.~\ref{eq:pose_update} (14) also leads to a performance decrease, showing the importance of them in refining pose and ultimately improving the depth learning.

For the choice of the feature map, we replace the online learned feature with different pretrained features, specifically the MASt3R~\cite{leroy2024grounding} feature (15) and the PixLoc feature (16). Experiment (15) outperforms the baseline, demonstrating that pose refinement is effective with a suitable visual feature metric. Still, our method (6) is better, demonstrating the necessity of learning robust visual features jointly during the training stage. In (16) with the PixLoc feature, it only reaches a lower depth accuracy than the baseline (a). This verifies the necessity of the current pose refinement design.

\begin{figure*}[t]
    \centering
    \includegraphics[width=1.0\textwidth]{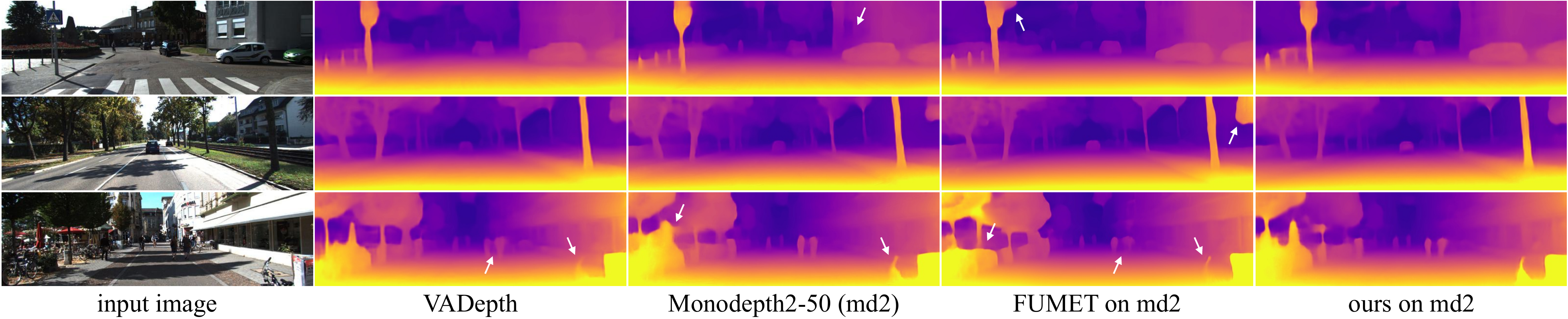}
    \vspace*{-1.5em}
    \caption{Qualitative comparison on KITTI Eigen~\cite{Geiger2013IJRR,eigen2015predicting} among scale-aware methods. White arrows highlight coarse wrong estimates.}
    \label{fig:qualitative_kitti_velocity}
\end{figure*}

\begin{figure*}[t]
    \centering
    \includegraphics[width=1.0\textwidth]{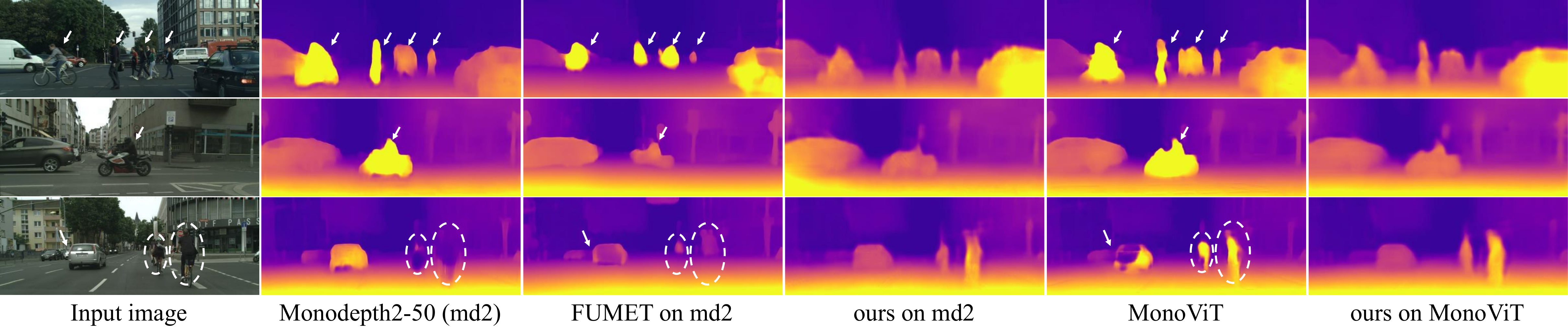}
    \vspace*{-1.5em}
    \caption{Qualitative comparison on Cityscapes~\cite{Cordts2016Cityscapes} among scale-aware methods. White arrows highlight coarse wrong estimates.}
    \label{fig:qualitative_cs_velocity}
    \vspace*{-1em}
\end{figure*}

\subsection{Qualitative Results}
\begin{figure}[h]
    \vspace{-0.5em}
    \centering
    \includegraphics[width=\columnwidth]{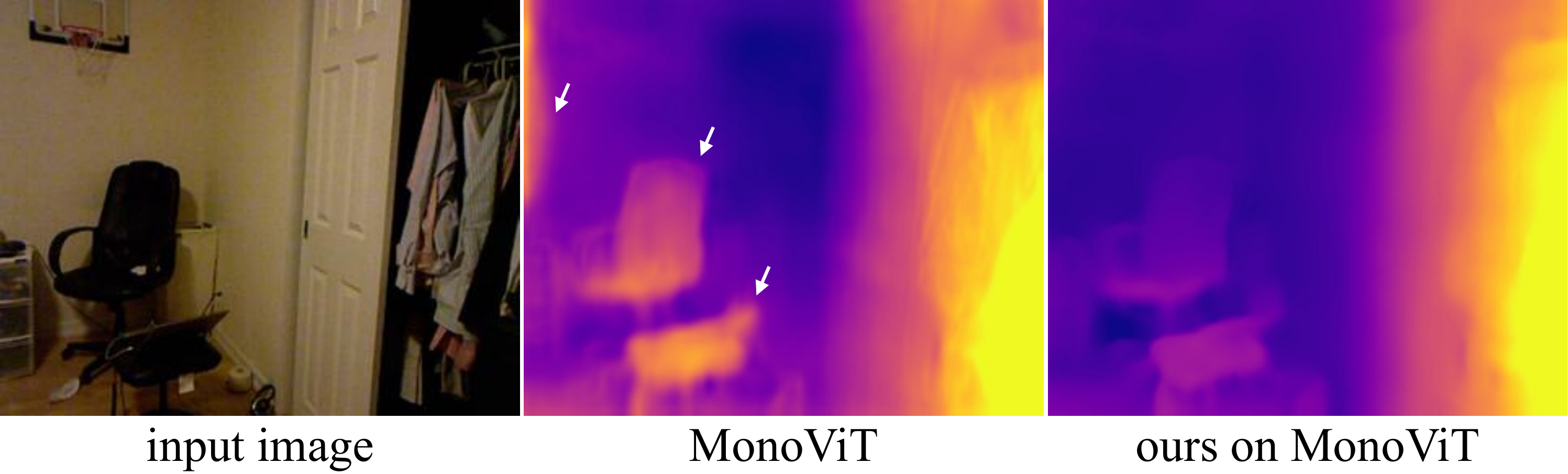}
    \caption{Qualitative comparison on NYUv2 \cite{Silberman:ECCV12}, errors marked in white.}
    \label{fig:qualitative_nyu}
    \vspace{-0.7em}
\end{figure}

In Fig.~\ref{fig:qualitative_kitti_velocity}, \ref{fig:qualitative_cs_velocity}, and \ref{fig:qualitative_nyu}, we show depth estimates on KITTI, Cityscapes, and NYUv2, respectively. Considering KITTI (Fig.~\ref{fig:qualitative_kitti_velocity}), FUMET~\cite{Kinoshita2023-hi} delivers similar outputs to the baseline md2-50, as in Table~\ref{tab:kitti_improved}. Paired with the same baseline, ours produces better depth estimates of thin objects, such as the delivery bicycle at the bottom-right of the lower image and the sun umbrella on the left of the same input.

Instead, on Cityscapes (Fig.~\ref{fig:qualitative_cs_velocity}), the gap between the methods increases, as seen in the quantitative results (Table~\ref{tab:cityscape_results}). Here, with more dynamic scenes, the baselines and FUMET~\cite{Kinoshita2023-hi} wrongly estimate the depth of many traffic participants, and FUMET even ignores the cyclists in the third image. On the other hand, thanks to aligned pose estimates during training, our method could significantly better understand the distance of traffic participants, albeit with blurrier outputs, which did not occur on KITTI.

In Fig.~\ref{fig:qualitative_nyu}, we show a qualitative comparison indoors on NYUv2~\cite{Silberman:ECCV12}. Here, the estimates from the up-to-scale baseline MonoViT~\cite{zhao2022monovit} are inaccurate for objects (e.g., chair and table) and exhibit high-frequency changes that are not present in the scene. Instead, applying ours on the same method resolves these issues, delivering superior estimates.
\begin{figure}[h]
    \centering    
    \includegraphics[width=\columnwidth]{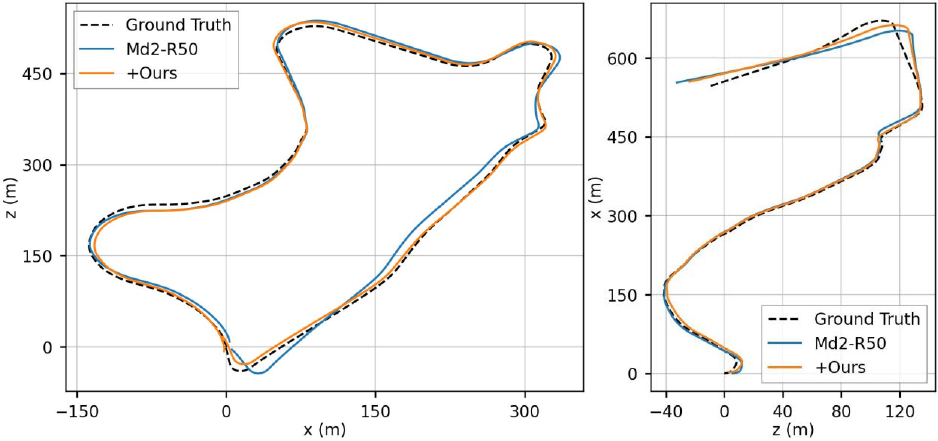}
    \caption{Visual odometry results on the KITTI Odometry Seq. 09 (Left) and Seq. 10 (Right), scaled as in Table~\ref{tab:kitti_odom}. The models are trained up-to-scale.}
    \label{fig:quali_odom}
\end{figure}
In Fig.~\ref{fig:quali_odom}, we show the visual odometry qualitative results on the KITTI Odometry~\cite{Geiger2012CVPR} sequences 09 and 10.  Our trajectory estimates align remarkably well with the straight segments, demonstrating outstanding pose estimation in the translation component, consistent with Table~\ref{tab:kitti_odom}. 

\textbf{Limitations and future work}
Our work focuses on improving the pose estimation, but does not alter the depth net. Thus, it could inherit issues from the baselines whenever a better pose cannot address them. The joint learning framework can be influenced by degraded depth estimates, leading to performance degradation, which also affects our method. Therefore, future work could focus on creating more robust frameworks.
While we see improvements on dynamic objects (Fig.~\ref{fig:qualitative_cs_velocity}), we do not explicitly modify networks to tackle them, which could lead to sub-optimal outputs in highly dynamic scenes. Future work could explore this to further improve the output reliability.
\section{Conclusion}
\label{sec:conclusion}


In this paper, we introduced SA4Depth, a novel approach to improve the pose-depth scale alignment and consistency within self-supervised monocular depth estimation. We achieve this by refining the pose estimate via feature reprojection. Compatible with any such self-supervised mono depth framework, our method leaves inference unchanged and delivers superior depth and pose estimates across various settings, making it a valid addition to self-supervised monocular depth estimators.







\bibliographystyle{IEEEtran}
\bibliography{IEEEabrv,main}

@STRING{IEEE_J_RO         = "{IEEE} Trans. Robot."}

@String(PAMI = {IEEE Trans. Pattern Anal. Mach. Intell.})

@String(CVPR= {IEEE Conf. Comput. Vis. Pattern Recog.})

@String(ICCV= {Int. Conf. Comput. Vis.})

@String(ECCV= {Eur. Conf. Comput. Vis.})

@String(NIPS= {Adv. Neural Inform. Process. Syst.})

@String(ICLR = {Int. Conf. Learn. Represent.})

@String(CSVT = {IEEE Trans. Circuit Syst. Video Technol.})

@String(PAMI  = {IEEE TPAMI})

@String(CVPR  = {CVPR})

@String(ICCV  = {ICCV})

@String(ECCV  = {ECCV})

@String(NIPS  = {NeurIPS})

@String(ICLR  = {ICLR})

@String(CSVT = {IEEE TCSVT})

@String(I3DV = {3DV})

@String(ICRA = {ICRA})

@String(RAL = {IEEE RAL})

@inproceedings{Kinoshita2023-hi,
  title={Camera Height Doesn’t Change: Unsupervised Training for Metric Monocular Road-Scene Depth Estimation},
  author={Kinoshita, Genki and Nishino, Ko},
  booktitle=ECCV,
  pages={57--73},
  year={2024},
  organization={Springer}
}

@inproceedings{guizilini2020packnet,
  title={{3D} packing for self-supervised monocular depth estimation},
  author={Guizilini, Vitor and Ambrus, Rares and Pillai, Sudeep and Raventos, Allan and Gaidon, Adrien},
  booktitle=CVPR,
  pages={2485--2494},
  year={2020}
}

@inproceedings{Yu2025-zo,
  title={Relative pose estimation through affine corrections of monocular depth priors},
  author={Yu, Yifan and Liu, Shaohui and Pautrat, R{\'e}mi and Pollefeys, Marc and Larsson, Viktor},
  booktitle=CVPR,
  pages={16706--16716},
  year={2025}
}

@inproceedings{Liu2024-wc,
  title={Mono-ViFI: A Unified Learning Framework for Self-supervised Single and Multi-frame Monocular Depth Estimation},
  author={Liu, Jinfeng and Kong, Lingtong and Li, Bo and Wang, Zerong and Gu, Hong and Chen, Jinwei},
  booktitle=ECCV,
  pages={90--107},
  year={2024},
  organization={Springer}
}

@inproceedings{Sarlin2021-ly,
  title={Back to the feature: Learning robust camera localization from pixels to pose},
  author={Sarlin, Paul-Edouard and Unagar, Ajaykumar and Larsson, Mans and Germain, Hugo and Toft, Carl and Larsson, Viktor and Pollefeys, Marc and Lepetit, Vincent and Hammarstrand, Lars and Kahl, Fredrik and others},
  booktitle=CVPR,
  pages={3247--3257},
  year={2021}
}

@inproceedings{Bangunharcana2023-iw,
  title={{DualRefine}: Self-supervised depth and pose estimation through iterative epipolar sampling and refinement toward equilibrium},
  author={Bangunharcana, Antyanta and Magd, Ahmed and Kim, Kyung-Soo},
  booktitle=CVPR,
  pages={726--738},
  year={2023}
}

@article{Xiang2022-bo,
  title={Visual attention-based self-supervised absolute depth estimation using geometric priors in autonomous driving},
  author={Xiang, Jie and Wang, Yun and An, Lifeng and Liu, Haiyang and Wang, Zijun and Liu, Jian},
  journal=RAL,
  volume={7},
  number={4},
  pages={11998--12005},
  year={2022},
  publisher={IEEE}
}

@inproceedings{zhao2022monovit,
  title={{MonoViT}: Self-supervised monocular depth estimation with a vision transformer},
  author={Zhao, Chaoqiang and Zhang, Youmin and Poggi, Matteo and Tosi, Fabio and Guo, Xianda and Zhu, Zheng and Huang, Guan and Tang, Yang and Mattoccia, Stefano},
  booktitle=I3DV,
  pages={668--678},
  year={2022},
  organization={IEEE}
}

@inproceedings{godard2019digging,
  title={Digging into self-supervised monocular depth estimation},
  author={Godard, Cl{\'e}ment and Mac Aodha, Oisin and Firman, Michael and Brostow, Gabriel J},
  booktitle=CVPR,
  pages={3828--3838},
  year={2019}
}

@inproceedings{Cordts2016Cityscapes,
title={The Cityscapes Dataset for Semantic Urban Scene Understanding},
author={Cordts, Marius and Omran, Mohamed and Ramos, Sebastian and Rehfeld, Timo and Enzweiler, Markus and Benenson, Rodrigo and Franke, Uwe and Roth, Stefan and Schiele, Bernt},
booktitle=CVPR,
year={2016}
}

@inproceedings{Geiger2012CVPR,
  author = {Andreas Geiger and Philip Lenz and Raquel Urtasun},
  title = {Are we ready for Autonomous Driving? The {KITTI} Vision Benchmark Suite},
  booktitle = CVPR,
  year = {2012}
}

@article{Geiger2013IJRR,
  author = {Andreas Geiger and Philip Lenz and Christoph Stiller and Raquel Urtasun},
  title = {Vision meets Robotics: The {KITTI} Dataset},
  journal = {IJRR},
  year = {2013}
}

@inproceedings{uhrig2017sparsity,
  title={Sparsity invariant cnns},
  author={Uhrig, Jonas and Schneider, Nick and Schneider, Lukas and Franke, Uwe and Brox, Thomas and Geiger, Andreas},
  booktitle=I3DV,
  pages={11--20},
  year={2017},
  organization={IEEE}
}

@inproceedings{eigen2015predicting,
  title={Predicting depth, surface normals and semantic labels with a common multi-scale convolutional architecture},
  author={Eigen, David and Fergus, Rob},
  booktitle=ICCV,
  pages={2650--2658},
  year={2015}
}

@inproceedings{watson2021temporal,
  title={The temporal opportunist: Self-supervised multi-frame monocular depth},
  author={Watson, Jamie and Mac Aodha, Oisin and Prisacariu, Victor and Brostow, Gabriel and Firman, Michael},
  booktitle=CVPR,
  pages={1164--1174},
  year={2021}
}

@inproceedings{zhou2017unsupervised,
  title={Unsupervised learning of depth and ego-motion from video},
  author={Zhou, Tinghui and Brown, Matthew and Snavely, Noah and Lowe, David G},
  booktitle=CVPR,
  pages={1851--1858},
  year={2017}
}

@inproceedings{zhan2018unsupervised,
  title={Unsupervised learning of monocular depth estimation and visual odometry with deep feature reconstruction},
  author={Zhan, Huangying and Garg, Ravi and Weerasekera, Chamara Saroj and Li, Kejie and Agarwal, Harsh and Reid, Ian},
  booktitle=CVPR,
  pages={340--349},
  year={2018}
}

@inproceedings{gasperini2023robust,
  title={Robust monocular depth estimation under challenging conditions},
  author={Gasperini, Stefano and Morbitzer, Nils and Jung, HyunJun and Navab, Nassir and Tombari, Federico},
  booktitle=CVPR,
  pages={8177--8186},
  year={2023}
}

@inproceedings{yan2021channel,
  title={Channel-wise attention-based network for self-supervised monocular depth estimation},
  author={Yan, Jiaxing and Zhao, Hong and Bu, Penghui and Jin, YuSheng},
  booktitle=I3DV,
  pages={464--473},
  year={2021},
  organization={IEEE}
}

@article{li2022monoindoor++,
  title={MonoIndoor++: Towards better practice of self-supervised monocular depth estimation for indoor environments},
  author={Li, Runze and Ji, Pan and Xu, Yi and Bhanu, Bir},
  journal=CSVT,
  volume={33},
  number={2},
  pages={830--846},
  year={2022},
  publisher={IEEE}
}

@article{bian2021auto,
  title={Auto-rectify network for unsupervised indoor depth estimation},
  author={Bian, Jia-Wang and Zhan, Huangying and Wang, Naiyan and Chin, Tat-Jun and Shen, Chunhua and Reid, Ian},
  journal=PAMI,
  volume={44},
  number={12},
  pages={9802--9813},
  year={2021},
  publisher={IEEE}
}

@inproceedings{he2016deep,
  title={Deep residual learning for image recognition},
  author={He, Kaiming and Zhang, Xiangyu and Ren, Shaoqing and Sun, Jian},
  booktitle=CVPR,
  pages={770--778},
  year={2016}
}

@inproceedings{von2020lm,
  title={{LM-Reloc}: {Levenberg-Marquardt} based direct visual relocalization},
  author={Von Stumberg, Lukas and Wenzel, Patrick and Yang, Nan and Cremers, Daniel},
  booktitle=I3DV,
  pages={968--977},
  year={2020},
  organization={IEEE}
}

@inproceedings{sarlin20superglue,
  title={Superglue: Learning feature matching with graph neural networks},
  author={Sarlin, Paul-Edouard and DeTone, Daniel and Malisiewicz, Tomasz and Rabinovich, Andrew},
  booktitle=CVPR,
  pages={4938--4947},
  year={2020}
}

@inproceedings{zhang2021holistic,
  title={Holistic 3d scene understanding from a single image with implicit representation},
  author={Zhang, Cheng and Cui, Zhaopeng and Zhang, Yinda and Zeng, Bing and Pollefeys, Marc and Liu, Shuaicheng},
  booktitle=CVPR,
  pages={8833--8842},
  year={2021}
}

@inproceedings{xu2024depthsplat,
  title={{DepthSplat}: Connecting Gaussian splatting and depth},
  author={Xu, Haofei and Peng, Songyou and Wang, Fangjinhua and Blum, Hermann and Barath, Daniel and Geiger, Andreas and Pollefeys, Marc},
  booktitle=CVPR,
  pages={16453--16463},
  year={2025}
}

@inproceedings{yang2024depth,
  title={Depth anything: Unleashing the power of large-scale unlabeled data},
  author={Yang, Lihe and Kang, Bingyi and Huang, Zilong and Xu, Xiaogang and Feng, Jiashi and Zhao, Hengshuang},
  booktitle=CVPR,
  pages={10371--10381},
  year={2024}
}

@inproceedings{bhat2021adabins,
  title={Adabins: Depth estimation using adaptive bins},
  author={Bhat, Shariq Farooq and Alhashim, Ibraheem and Wonka, Peter},
  booktitle=CVPR,
  pages={4009--4018},
  year={2021}
}

@inproceedings{gasperini2021r4dyn,
  title={R4dyn: Exploring radar for self-supervised monocular depth estimation of dynamic scenes},
  author={Gasperini, Stefano and Koch, Patrick and Dallabetta, Vinzenz and Navab, Nassir and Busam, Benjamin and Tombari, Federico},
  booktitle=I3DV,
  pages={751--760},
  year={2021},
  organization={IEEE}
}

@inproceedings{Silberman:ECCV12,
  author    = {Nathan Silberman, Derek Hoiem, Pushmeet Kohli and Rob Fergus},
  title     = {Indoor Segmentation and Support Inference from RGBD Images},
  booktitle = ECCV,
  year      = {2012}
}

@inproceedings{zhao2020towards,
  title={Towards better generalization: Joint depth-pose learning without posenet},
  author={Zhao, Wang and Liu, Shaohui and Shu, Yezhi and Liu, Yong-Jin},
  booktitle=CVPR,
  pages={9151--9161},
  year={2020}
}

@inproceedings{zhan2020visual,
  title={Visual odometry revisited: What should be learnt?},
  author={Zhan, Huangying and Weerasekera, Chamara Saroj and Bian, Jia-Wang and Reid, Ian},
  booktitle=ICRA,
  pages={4203--4210},
  year={2020},
  organization={IEEE}
}

@inproceedings{wang2018learning,
  title={Learning depth from monocular videos using direct methods},
  author={Wang, Chaoyang and Buenaposada, Jos{\'e} Miguel and Zhu, Rui and Lucey, Simon},
  booktitle=CVPR,
  pages={2022--2030},
  year={2018}
}

@article{bian2019unsupervised,
  title={Unsupervised scale-consistent depth and ego-motion learning from monocular video},
  author={Bian, Jiawang and Li, Zhichao and Wang, Naiyan and Zhan, Huangying and Shen, Chunhua and Cheng, Ming-Ming and Reid, Ian},
  journal=NIPS,
  volume={32},
  year={2019}
}

@article{mur2017orb,
  title={Orb-slam2: An open-source slam system for monocular, stereo, and rgb-d cameras},
  author={Mur-Artal, Raul and Tard{\'o}s, Juan D},
  journal=IEEE_J_RO,
  volume={33},
  number={5},
  pages={1255--1262},
  year={2017},
  publisher={IEEE}
}

@inproceedings{leroy2024grounding,
  title={Grounding image matching in 3d with mast3r},
  author={Leroy, Vincent and Cabon, Yohann and Revaud, J{\'e}r{\^o}me},
  booktitle=ECCV,
  pages={71--91},
  year={2024},
  organization={Springer}
}

@inproceedings{kingma2014adam,
  author    = {Kingma, Diederik P. and Ba, Jimmy},
  title     = {Adam: A Method for Stochastic Optimization},
  booktitle = {ICLR},
  year      = {2015},
  url       = {https://arxiv.org/abs/1412.6980}
}

@article{cabon2020virtual,
  title={Virtual kitti 2},
  author={Cabon, Yohann and Murray, Naila and Humenberger, Martin},
  journal={arXiv preprint arXiv:2001.10773},
  year={2020}
}

@inproceedings{zhang2025hybrid,
  title={Hybrid-grained feature aggregation with coarse-to-fine language guidance for self-supervised monocular depth estimation},
  author={Zhang, Wenyao and Liu, Hongsi and Li, Bohan and He, Jiawei and Qi, Zekun and Wang, Yunnan and Zhao, Shengyang and Yu, Xinqiang and Zeng, Wenjun and Jin, Xin},
  booktitle=ICCV,
  pages={6678--6692},
  year={2025}
}

@inproceedings{hu2025depthcrafter,
  title={Depthcrafter: Generating consistent long depth sequences for open-world videos},
  author={Hu, Wenbo and Gao, Xiangjun and Li, Xiaoyu and Zhao, Sijie and Cun, Xiaodong and Zhang, Yong and Quan, Long and Shan, Ying},
  booktitle=CVPR,
  pages={2005--2015},
  year={2025}
}

@inproceedings{chen2025video,
  title={Video depth anything: Consistent depth estimation for super-long videos},
  author={Chen, Sili and Guo, Hengkai and Zhu, Shengnan and Zhang, Feihu and Huang, Zilong and Feng, Jiashi and Kang, Bingyi},
  booktitle=CVPR,
  pages={22831--22840},
  year={2025}
}

@inproceedings{tateno2017cnn,
  title={Cnn-slam: Real-time dense monocular slam with learned depth prediction},
  author={Tateno, Keisuke and Tombari, Federico and Laina, Iro and Navab, Nassir},
  booktitle=CVPR,
  pages={6243--6252},
  year={2017}
}

\newpage
\clearpage
\appendix

\section*{A. Robustness against Depth Estimation Noise}
\label{app:robust}
We present quantitative results of new experiments in Tab.~\ref{tab:robust} to investigate the robustness of the joint training framework against degraded depth. The experiments are conducted with the baseline Monodepth2 with ResNet-18 (Md2-18) on the KITTI~\cite{Geiger2013IJRR} dataset. In the experiments, we inject relative pixel-wise noise into the depth estimates as $\epsilon \sim \mathcal{N}(0, \sigma^2)$, i.e., following a Gaussian distribution, while keeping the other parts of the joint training framework unchanged. The degraded depth predictions are
\begin{equation}
    D_{t} = D_{t} \odot (1 + \epsilon),
\end{equation}
where the $\epsilon$ is of the same shape as depth estimates $D_{t}$ and $\odot$ is an element-wise multiplication.

\begin{table}[h]
\centering
\caption{Analysis on the robustness against degraded depth estimation, building upon Md2-18 trained on KITTI Eigen split. Noise Std $\sigma$ is the \textit{std} of the Gaussian noise applied on the depth estimates. Evaluated with GT median scaling. \textbf{Best}.}
\label{tab:robust}
\begin{tabular}{ll|ccc}
    \toprule 
    Noise Std $\sigma$ & Model & Abs Rel & $\delta < 1.25$ \\
    \midrule 
    0 (as in Tab.~\ref{tab:kitti_improved})  & Md2-18      & 0.090 & 0.914 \\
    0 (as in Tab.~\ref{tab:kitti_improved})  & + ours     & \textbf{0.087} & \textbf{0.918} \\
    \midrule
    0.05  & Md2-18    & 0.099 & 0.904 \\
    0.05  & + ours   & \textbf{0.096} & \textbf{0.907} \\
    \midrule
    0.1  & Md2-18    & 0.114 & 0.879 \\
    0.1  & + ours   & \textbf{0.110} & \textbf{0.886} \\
    \bottomrule
\end{tabular}
\end{table}

From the results, we observe a performance drop in the baseline model as the noise level increases with larger noise std $\sigma$. However, when our method is applied on top of the baseline, performance improves across all settings, demonstrating that it consistently outperforms the baseline. Nevertheless, our approach inherits the limitations of SSMDE joint learning, in which degraded depth during training negatively affects final performance. Because our pose refinement uses the learned depth prediction as scene geometry to optimize the pose prediction and same as the baseline joint learning method Monodepth2\cite{godard2019digging} (Md2), when the scene geometry degrades, the feature correspondences built via reprojection become invalid, resulting in degraded performance. However, in a normal training process, without reinforced depth noise, the joint learning can learn both depth and pose effectively, even though depth or pose at the intermediate steps can be noisy, as shown in the Tab.~\ref{tab:robust} in the rows with $\sigma=0$.

\section*{B: Video Depth Evaluation for Scale Consistency Comparison}
\label{app:video_depth}

\begin{table}[h]
    \caption{Video depth on KITTI sequences of 110 frames. \\seq. scale std. = $std(\frac{scale}{mean(scale)})$.\\GT: using GT depth for supervision. \textbf{Best}.}
    \label{tab:video_depth}
    \centering
    \resizebox{\columnwidth}{!}{%
    \begin{tabular}{ll|c|cccc}
    \toprule
    Method & GT & seq. scale std. & Abs Rel & RMSE & $\delta < 1.25$ \\
    \midrule
        DepthCrafter~\cite{hu2025depthcrafter} & \checkmark & 0.193
        & 0.103 & 3.963 & 0.892  \\
        VDA\cite{chen2025video} & \checkmark & 0.083
        & 0.082 & \textbf{3.318} & \textbf{0.947} \\
    \midrule
        MonoViT    & $\times$ & 0.093     
        & 0.083 & 3.713 & 0.932\\
        \cellcolor{oursblue}MonoViT \textbf{+ ours}         
        & \cellcolor{oursblue}$\times$
        & \cellcolor{oursblue}\textbf{0.055}       
        & \cellcolor{oursblue}\textbf{0.078}       
        & \cellcolor{oursblue}3.553 
        & \cellcolor{oursblue}0.942 
        \\
    \bottomrule
    \end{tabular}
    }
\end{table}

We evaluate here the video depth estimation performance of our model paired with MonoViT~\cite{zhao2022monovit} using the DepthCrafter~\cite{hu2025depthcrafter} evaluation setup on the KITTI dataset validation set (13 scenes, each of length 110 frames), as shown in Tab.~\ref{tab:video_depth}. Following DepthCrafter, the depth estimates are aligned with the ground truth using a scale and shift for each sequence. According to the evaluation metrics, our model is on par with the current SOTA method Video-Depth-Anything~\cite{chen2025video} (shown as VDA in Tab.~\ref{tab:video_depth}), which is trained with ground truth depth supervision, while our method is trained self-supervised. This shows that the current self-supervised methods are comparable to supervised methods on KITTI, with improvements in scale consistency when our method is applied. However, we acknowledge that there may be a generalization gap, as video depth methods are trained on large-scale mixed datasets, and the SSMDE models, including ours, are trained on a single dataset.

The sequence scale std is calculated by 
\begin{equation}
    \text{Seq. Scale Std} = std(\frac{[s_0, ..., s_n]}{mean([s_0, ..., s_n])}),
\end{equation}
where $s_i$ is the scaling factor for a single sequence to align with the ground truth, and the lower scale standard deviation shows better depth estimation scale consistency. Due to the up-to-scale depth estimates and the large scaling difference between models, we calculate the standard deviation of the mean-normalized per-sequence scaling factor. As shown in Tab.~\ref{tab:video_depth}, when paired with MonoViT~\cite{zhao2022monovit}, our method outperforms SOTA supervised video depth estimation methods in scale consistency and improves upon the baseline, demonstrating good scale stability across different environments.

\section*{C: Internalization of SLAM}
\label{app:slam}

SLAM approaches, such as CNN-SLAM~\cite{tateno2017cnn}, incorporate learnable depth networks to maintain scale consistency during global map alignment.
While CNN-SLAM fuses depth predictions from a supervised network into a monocular SLAM pipeline to leverage learned dense depth and absolute scale, our work inherits the fundamental goal of ensuring geometric consistency while advancing the paradigm by internalizing this process. Specifically, within our self-supervised framework, we embed SLAM-style optimization for keyframe pose alignment directly into the training stage. By aligning adjacent frames via feature reprojection based on current depth estimates, our method dynamically enforces scale consistency between depth and pose, thereby fundamentally enhancing the network's consistent geometric understanding. Since pose refinement is only performed during training, there's no additional inference cost for our method.

\section*{D: Extra Qualitative Results}
\label{app:sup_quali}

\begin{figure*}[ht]
    \centering
    \subfloat[Qualitative results on KITTI. White arrows highlight coarse wrong estimates.]{%
        \includegraphics[width=\textwidth]{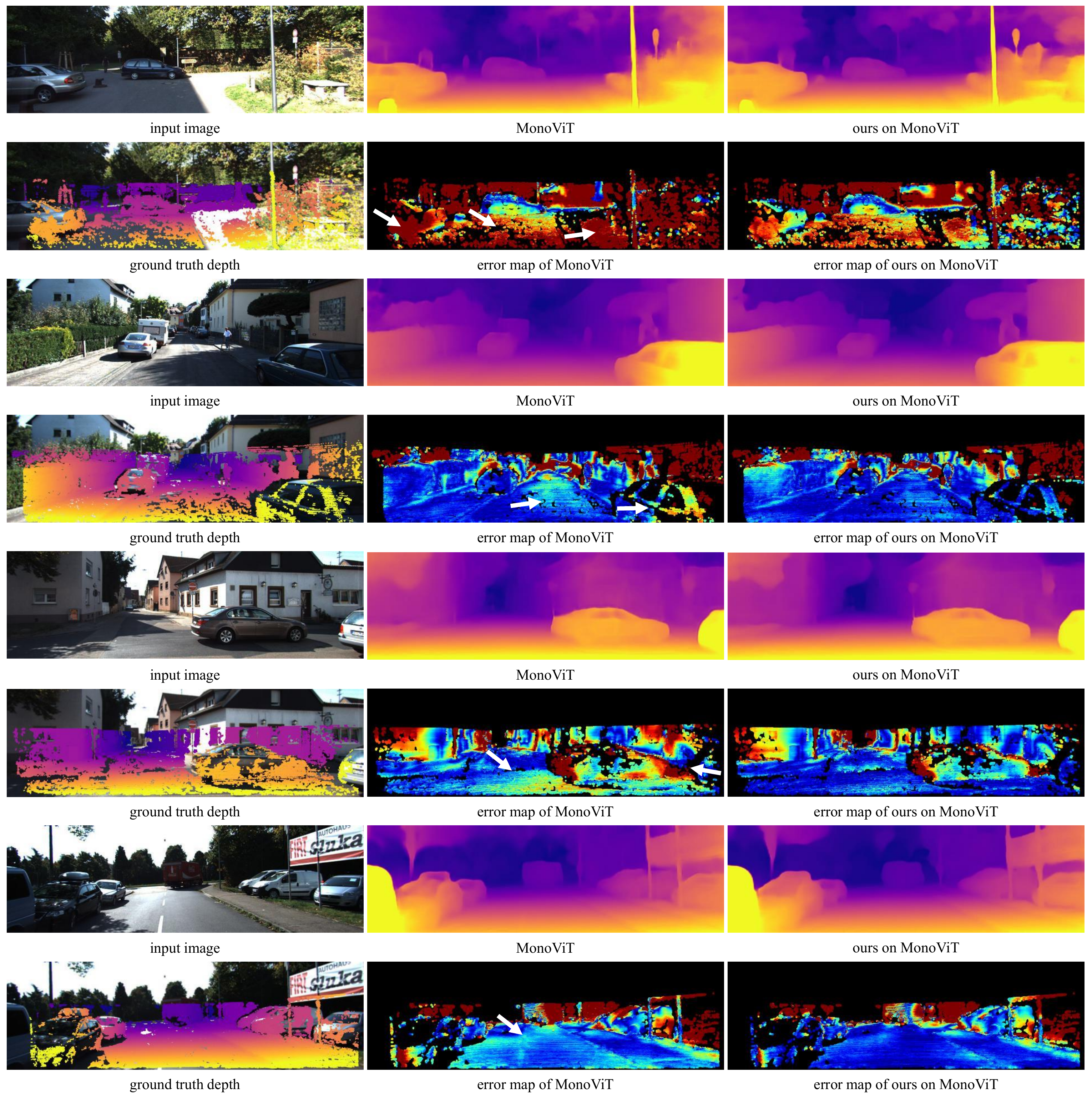}%
    }
    \vspace{0.5em} 
    
    \subfloat[Abs Rel error map colorbar.]{%
        \includegraphics[width=0.5\textwidth]{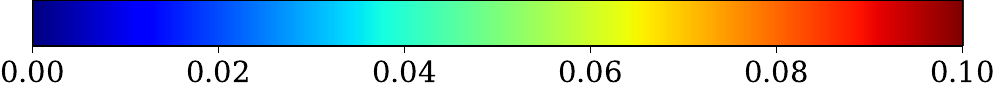}%
    }
    \caption{\label{fig:sup_kitti} More qualitative comparison on KITTI Eigen Split. The depth predictions are up-to-scale. Comparison also includes every second row, the improved ground truth, and the error maps showing the magnitude of the error metric Abs Rel. Improved areas are highlighted with arrows in the error maps.}    
\end{figure*}

\textbf{Depth on KITTI}
We show more qualitative results in Figure~\ref{fig:sup_kitti} using MonoViT~\cite{zhao2022monovit} as baseline, where the depth estimates are up-to-scale. We include corresponding ground truth depth and error maps to stress on the improved areas. The error maps show the depth accuracy metric Abs Rel error, calculated as:
\begin{equation}
    \text{Abs Rel} = \left|\frac{D_{Pred}- D_{GT}}{D_{GT}}\right|,
    \label{eq:sup_abs_rel}
\end{equation}

where the depth predictions $D_{Pred}$ are median scaled by the scale ratio $median(D_{GT})/median(D_{Pred})$ and for the ground truth $D_{GT}$ we use the improved ground truth from KITTI dataset.
Noticeably, our method remarkably improves the depth estimates on the vehicle surface (in all groups), ground (in the $2^{nd}$, $3^{rd}$, and $4^{th}$ groups), and vegetation (in the $1^{st}$ group). Thanks to the scale-aligned pose during training through our method, training losses can reflect the delicate depth errors in these areas, and the depth net can effectively learn their geometry.

\begin{figure*}[htbp]
    \centering
    \includegraphics[width=\textwidth]{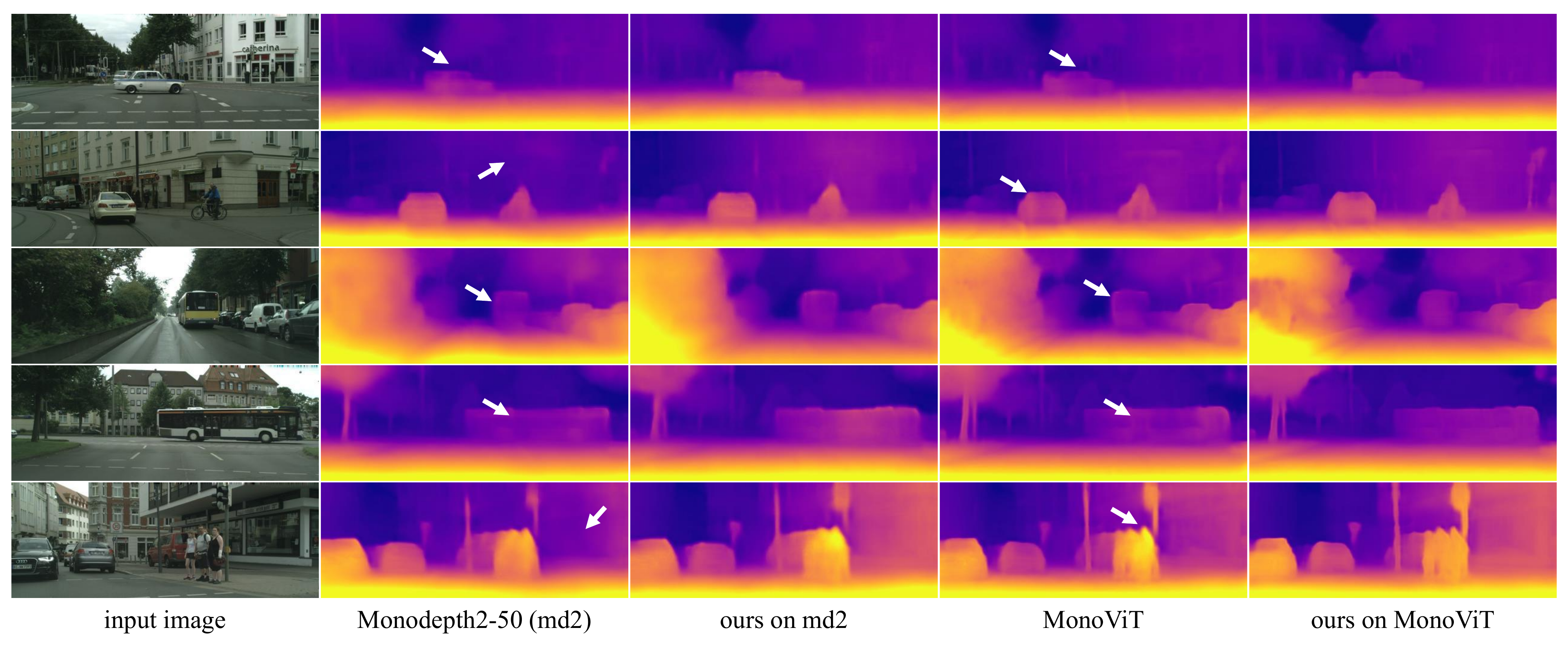}
    \caption{\label{fig:sup_cityscapes}More qualitative comparison on Cityscapes. The depth estimates are up-to-scale. White arrows highlight coarse wrong estimates.}    
\end{figure*}

\textbf{Depth on Cityscapes}
We include more qualitative comparisons on Cityscapes in Figure~\ref{fig:sup_cityscapes}, where we compare qualitative results among Monodepth2-ResNet50, MonoViT, and ours applied to them. The results are up-to-scale. Remarkably, Monodepth2 exhibits wrong estimates on background buildings (rows 2 and 5). Thanks to an aligned, consistent scale, our method improves estimates in such background regions. Both baselines show mistakes on moving vehicles (rows 1, 3, 4) where holes appear on the vehicles' surface, and our method alleviates these errors. In row 5, MonoViT shows erroneous estimates of the pedestrians, while our method estimates them correctly.

\begin{figure*}[htbp]
    \centering
    \includegraphics[width=\textwidth]{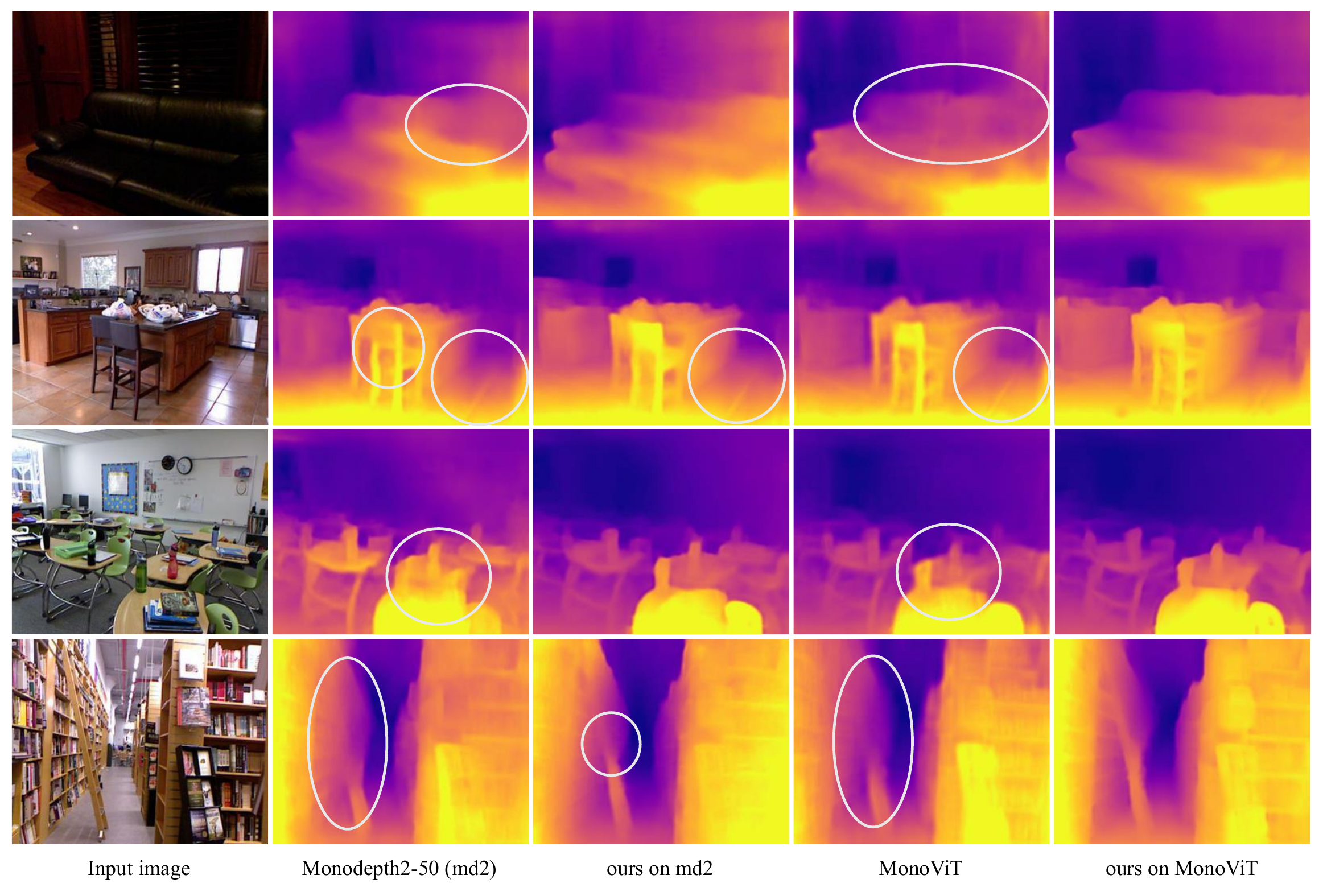}
    \caption{More qualitative comparison on NYUv2. The depth estimates are up-to-scale. White marks highlight coarse wrong estimates.}
    \label{fig:sup_nyu}
\end{figure*}

\textbf{Depth on NYUv2}
We show more qualitative results on NYUv2 in Figure~\ref{fig:sup_nyu}, pairing our method with baselines Monodepth2-ResNet50 and MonoViT. Noticeably, the baselines have difficulties in estimating the depth of the sofa (row 1), tables (row 3), and the thin ladder (row 4). Our method estimates them correctly, reflecting the correct geometry. In row 2, Monodepth2 ignores the backrest of a chair, and both baselines struggle with the ground with reflections. Again, applying our method mitigates the errors.

\section*{E: Feature and Confidence Maps}
\label{app:sub_feat_conf_quali}

\begin{figure*}[t]
    \centering
    \subfloat[Feature and confidence maps.]{%
        \includegraphics[width=\textwidth]{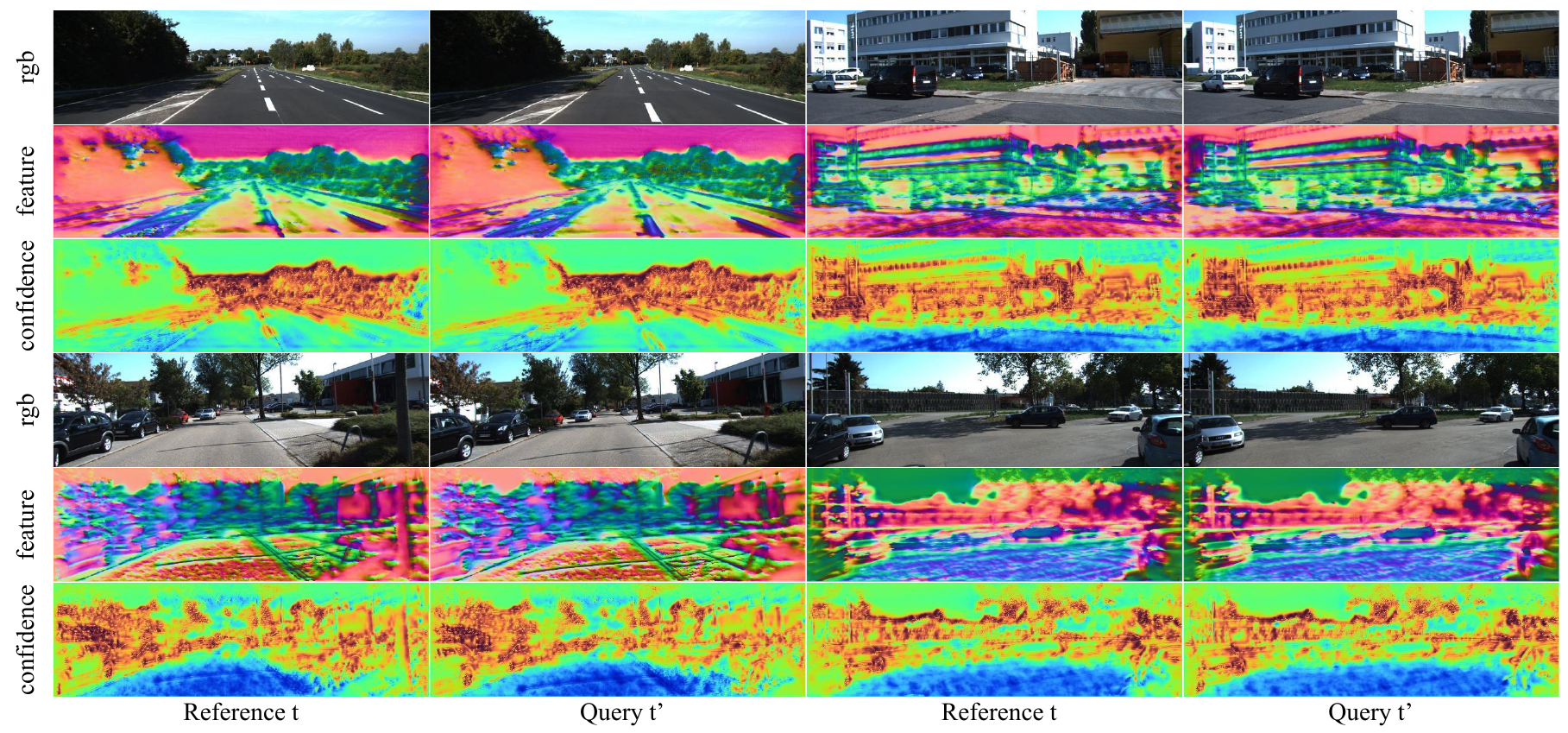}%
    }
    \vspace{0.5em} 
    
    \subfloat[Confidence colorbar.]{%
        \includegraphics[width=0.5\textwidth]{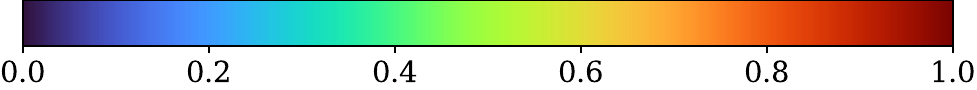}%
    }
    \caption{\label{fig:feature_conf} Feature and confidence maps extracted by the feature net, which is trained on KITTI. The feature maps are processed by PCA. The confidence maps weight corresponding feature residuals in the pose refinement cost function, where high confidence indicates the significance of the corresponding area in refining the camera pose.}
\end{figure*}

We show the feature and confidence maps extracted for our pose refinement trained on the KITTI dataset in Figure~\ref{fig:feature_conf}. The feature maps are processed by Principal Component Analysis (PCA). By comparing the maps for reference frame t and the adjacent query frame t', we can observe the feature consistency on the same object in the feature map of both frames. The consistent feature is essential for pose refinement. Also, the confidence maps are used to mitigate the influence of textureless regions. The shown confidence maps emphasize (with red) structured areas, such as lane markings, buildings, trees, and parked vehicles, for robust pose refinement, while having a lower value (with blue) in textureless areas, such as road and sky, showing the different weight for the pose refinement cost calculation.

\end{document}